\documentclass[10pt,twocolumn,letterpaper]{article}

\usepackage[pagenumbers]{cvpr}

\usepackage{graphicx}
\usepackage{amsmath}
\usepackage{amssymb}
\usepackage{booktabs}

\usepackage{textgreek}
\usepackage{siunitx}
\usepackage{algorithm}
\usepackage[noend]{algpseudocode}
\usepackage{listings}

\usepackage[pagebackref,breaklinks,colorlinks]{hyperref}

\usepackage[capitalize]{cleveref}
\crefname{section}{Sec.}{Secs.}
\Crefname{section}{Section}{Sections}
\Crefname{table}{Table}{Tables}
\crefname{table}{Tab.}{Tabs.}

\def\cvprPaperID{*****}
\def\confName{CVPR}
\def\confYear{2023}

\begin{document}

\title{EduceLab-Scrolls: Verifiable Recovery of Text from Herculaneum Papyri using X-ray CT}

\author{Stephen Parsons\\
{\tt\small stephen.parsons@uky.edu}\\
\and
C. Seth Parker\\
\and
Christy Chapman\\
\and
Mami Hayashida\\
\and
W. Brent Seales\\ \\
University of Kentucky\\
Lexington, KY, USA\\
}
\maketitle

\begin{abstract}
We present a complete software pipeline for revealing the hidden texts of the Herculaneum papyri using X-ray CT images.
This enhanced virtual unwrapping pipeline combines machine learning with a novel geometric framework linking 3D and 2D images.
We also present EduceLab-Scrolls, a comprehensive open dataset representing two decades of research effort on this problem.
EduceLab-Scrolls contains a set of volumetric X-ray CT images of both small fragments and intact, rolled scrolls.
The dataset also contains 2D image labels that are used in the supervised training of an ink detection model.
Labeling is enabled by aligning spectral photography of scroll fragments with X-ray CT images of the same fragments, thus creating a machine-learnable mapping between image spaces and modalities.
This alignment permits supervised learning for the detection of carbon ink in X-ray CT, a task that is has been considered impossible even for human expert labelers.
To our knowledge, this is the first aligned dataset of its kind and is the largest dataset ever released in the heritage domain.
Our method is capable of revealing accurate lines of text on scroll fragments with known ground truth.
Revealed text is verified using visual confirmation, quantitative image metrics, and scholarly review.
EduceLab-Scrolls has also enabled the discovery, for the first time, of hidden texts from the Herculaneum papyri, which we present here.
We anticipate that the EduceLab-Scrolls dataset will generate more textual discovery as research continues.
\end{abstract}

\section{Introduction} \label{sec:intro}

In 79 AD, the eruption of Mount Vesuvius destroyed the ancient Roman towns of Pompeii and Herculaneum, bombarding Pompeii with volcanic rock and debris and burying Herculaneum in up to 25 meters of pyroclastic ash.
Among the buried structures in Herculaneum was a large seaside villa that, upon its discovery and excavation in 1752-54, was found to contain a large library of papyrus rolls.
This collection of more than 900 scrolls was damaged yet remarkably preserved.
The fast-moving river of hot gas and ash from Vesuvius had simultaneously heated the papyrus scrolls to combustion temperature and extinguished any oxygen around them, carbonizing them in place.
Blackened and brittle like charcoal, the Herculaneum scrolls were rendered highly fragile, easily falling to pieces with light contact.

Referred to as the ``Herculaneum papyri,'' this collection represents the only library from antiquity ever discovered in situ, prompting various attempts over the years to physically unroll them and reveal their contents.
These efforts have generated a substantial quantity of legible text and scholarship, but have also led to widespread destruction.
Many scrolls that were opened were broken into many small fragments, or once had portions that are now lost entirely.
Hundreds of other scrolls were left intact, their contents completely unknown to this day.

In recent decades, advanced imaging technologies have introduced counter-methods to the physically destructive approaches of the past.
In particular, X-ray micro-computed tomography (micro-CT, or CT) has proven successful as a tool for noninvasive textual discovery \cite{seales2016engedi}.
However, despite the successful application of high-tech approaches to similar collections around the world, noninvasive techniques struggle to surface the contents of the Herculaneum scrolls due to several technical challenges.
First, the detection of ink on the writing substrate is problematic.
Known as ``lamp black'' and made from fine soot, the ink used by the Herculaneum writers is almost pure carbon.
The papyrus writing substrate is organic plant material, also largely made of carbon.
The result is little to no visible ink contrast in X-ray, which is key to the prior success of virtual unwrapping methods.
Images generated by the traditional virtual unwrapping of Herculaneum CT data show a blank page with no writing.

Another primary challenge concerns the physical interior structure of the scrolls -- multi-layered, wrapped, compressed, and highly irregular -- and the three-dimensional image segmentation problem it generates.
To manufacture papyrus, two or more layers of plant fibers are pressed together and naturally adhere, forming a multi-layered single sheet.
The extreme environment of carbonization often causes these sheets to bubble or delaminate, increasing the number of apparent papyrus layers in CT images.
Further, the scrolls are large, contain many wraps (on the order of hundreds), have adjacent wraps that are pressed tightly together, and, in many cases, have been compressed and crumpled during their burial.
Micro-CT is able to penetrate to the scroll interior at high enough spatial resolutions to clearly show this internal structure of warped layers, and even the detailed grid-like structure of the papyrus sheet fibers.
But deciphering these forms into the specific surfaces on which the ink lies is nonetheless difficult.

Though formidable, we believe these challenges can now be addressed as the result of multiple advancements.
Imaging resolution, machine learning tools, and computational methods are simultaneously maturing enough to enable strong results.
With the right experimental framework, the low contrast ink can be detected using machine learning models.
This method leverages the otherwise destructive physical unrolling efforts, as the detached fragments with visible ink can be used as a training dataset.
The segmentation problem is also addressable, as improved methods that follow the structure of the papyrus fibers are able to extract order from the chaos of the damaged material.

To this end, we present EduceLab-Scrolls, a multimodal image dataset of Herculaneum papyri, along with a machine learning-based method for ink detection.
EduceLab-Scrolls represents two decades of research progress into the noninvasive discovery of the hidden texts of the Herculaneum scrolls.
The dataset contains high resolution volumetric images from X-ray micro-CT of both intact, rolled scrolls and small detached scroll fragments.
To our knowledge, this is the largest such dataset released in the heritage domain.

A primary contribution of EduceLab-Scrolls is not only the quantity and resolution of the CT images, but also the aligned image labels generated using a multi-step process described in Appendix \ref{methods}.
The volumetric X-ray CT images are aligned with 2D surface images, such as full color and spectral photographs, of the same objects.
These aligned labels enable supervised learning for micro-CT input images, leading to trained models capable of detecting carbon ink in CT.
To our knowledge, this image alignment framework is unique to this work, and this is the largest release of such an aligned image dataset.

We also present a baseline method for learning the CT $\rightarrow$ ink-presence mapping, a refinement built on top of previous work \cite{parker2019inkid}.
This method is trained on EduceLab-Scrolls and is capable of recovering hidden text (shown in Figure \ref{fig:fragments-ours-hidden}) given only X-ray CT images as input.
This method operates only on image pixels, identifying them as ink or not ink.
The method and its trained model possess no knowledge of character alphabets, optical character recognition (OCR), paleography, or handwriting analysis.
Thus it is completely incapable of recognizing letter characters or shapes.
The text characters that emerge result purely from plotting the local ink detection ``spots'' across a generated image.

This noninvasive recovery of hidden Herculaneum text, sealed inside the scroll since its destruction in 79 AD, is a world first on a research problem that has been pursued since the scrolls were first unearthed.
Despite other claims to noninvasive textual discovery from Herculaneum papyri, we assert this is the first method capable of generating substantial text suitable for literary scholarship.
To be readable, letters should form lines with uniform and expected scale, spacing, and margins \cite{60minutes}.
Our method is the first capable of meeting this threshold, and further is the first to verify readability using scroll fragments with known ground truth.
We evaluate our method against this threshold with visual confirmation, quantitative image metrics, and by image transcriptions from Herculaneum papyrus scholars.

With EduceLab-Scrolls, we invite the scholarly and technical community to validate and extend these results.
Though the dataset and method presented are designed to fulfill the specific applied objective of reading Herculaneum papyri, we believe they also offer a new path forward for other domains and detection problems.
For example, supervised learning on medical X-ray CT often uses human expert labeling: a doctor examines the CT manually, and labels what is visually observed.
A model is then trained to reproduce this detection.
This work suggests a new paradigm.
We demonstrate that X-ray micro-CT captures information that is not visible to the human eye, and that models can be trained to detect such signals.
To do this, images from a different modality containing the sought after information are aligned with the CT data and used as a training dataset.
By leveraging the correspondence between imaging methods, previously invisible signals become visible.

\section{Results}

\addtolength{\tabcolsep}{12pt}   
\begin{figure*}
    \centering
    \captionsetup{justification=centering}

    \begin{subfigure}{0\textwidth}
        \refstepcounter{subfigure}\label{fig:fragments-ground-truth}
    \end{subfigure}
    \begin{subfigure}{0\textwidth}
        \refstepcounter{subfigure}\label{fig:fragments-ours-surface}
    \end{subfigure}
    \begin{subfigure}{0\textwidth}
        \refstepcounter{subfigure}\label{fig:fragments-ours-hidden}
    \end{subfigure}

    \begin{tabular}{p{0.16\textwidth} p{0.16\textwidth} p{0.16\textwidth} l}
        \includegraphics[width=0.16\textwidth]{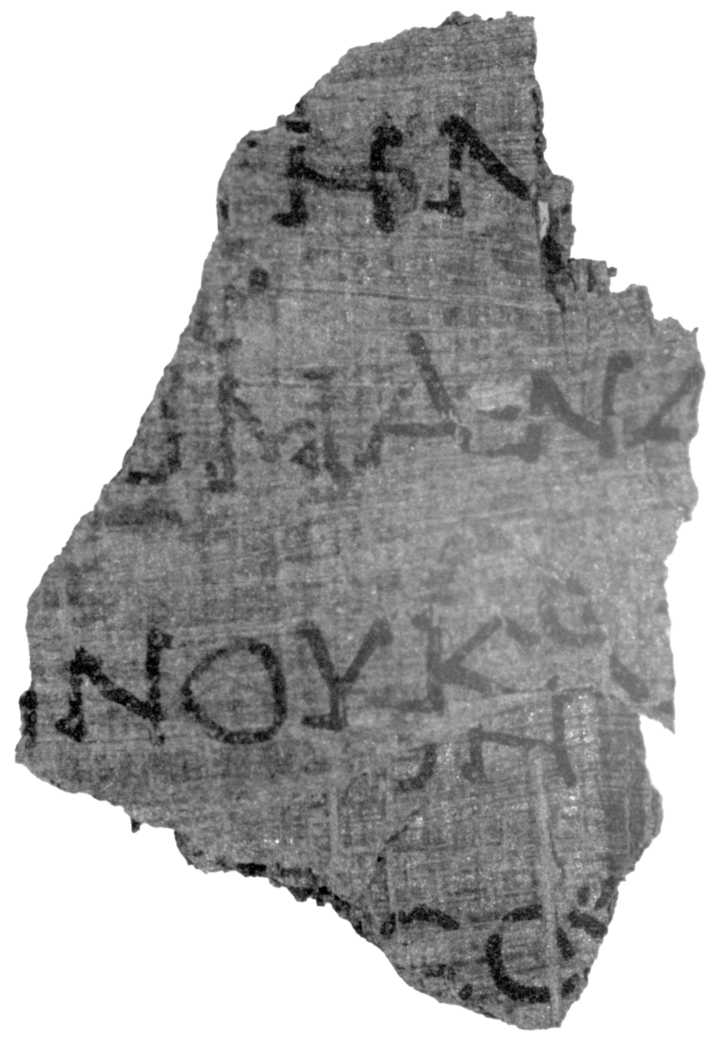} & 
        \includegraphics[width=0.16\textwidth]{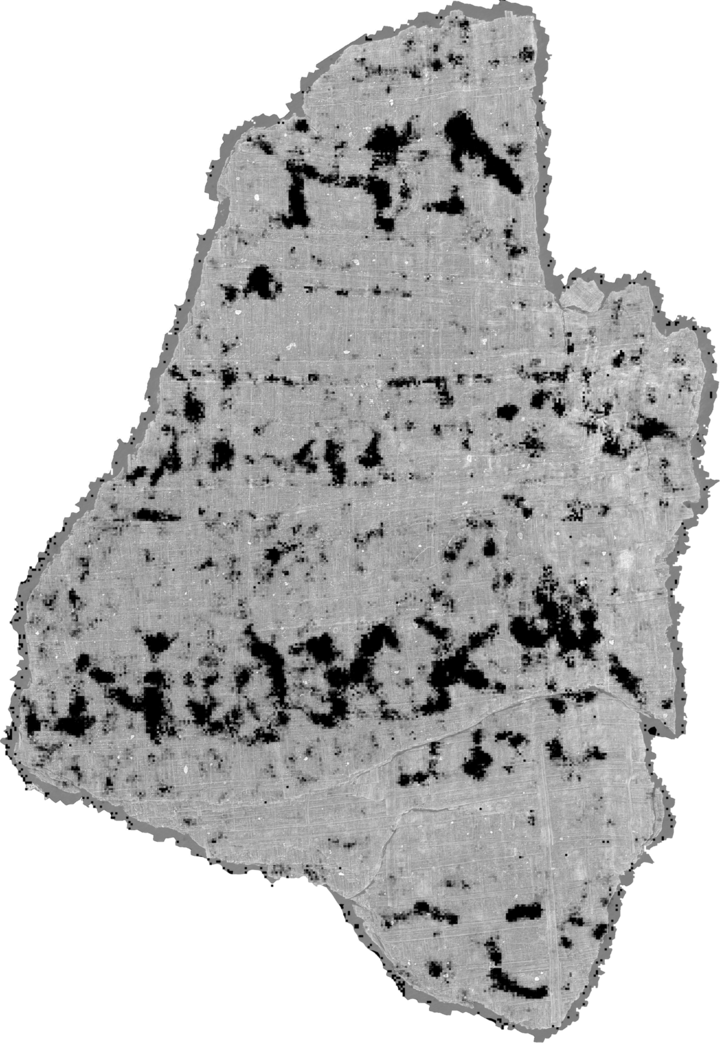} & \\
        \\

        \includegraphics[width=0.16\textwidth]{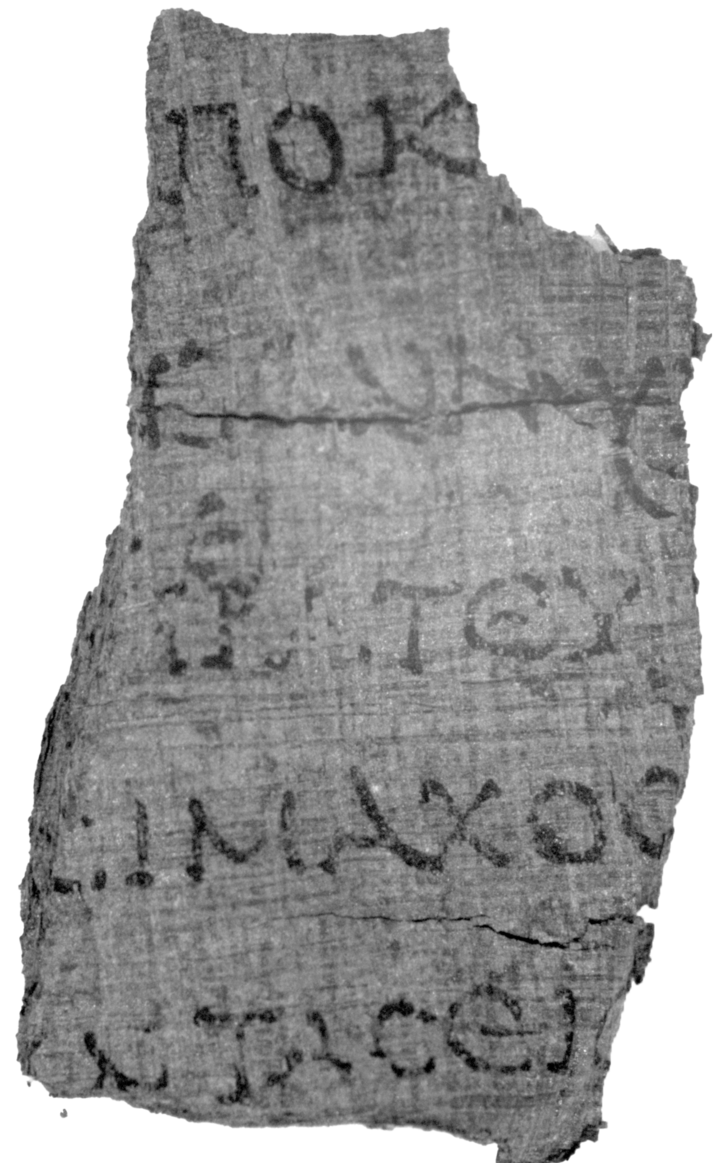} &
        \includegraphics[width=0.16\textwidth]{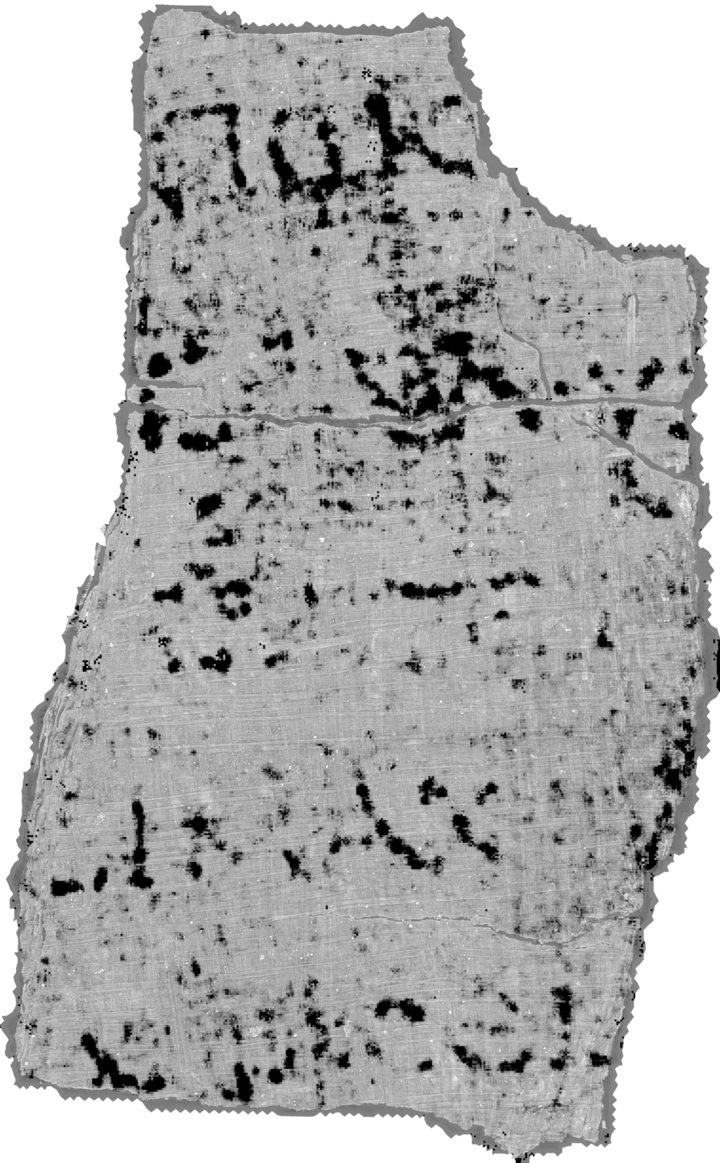} &
        \raisebox{0.3\height}{\includegraphics[width=0.14\textwidth]{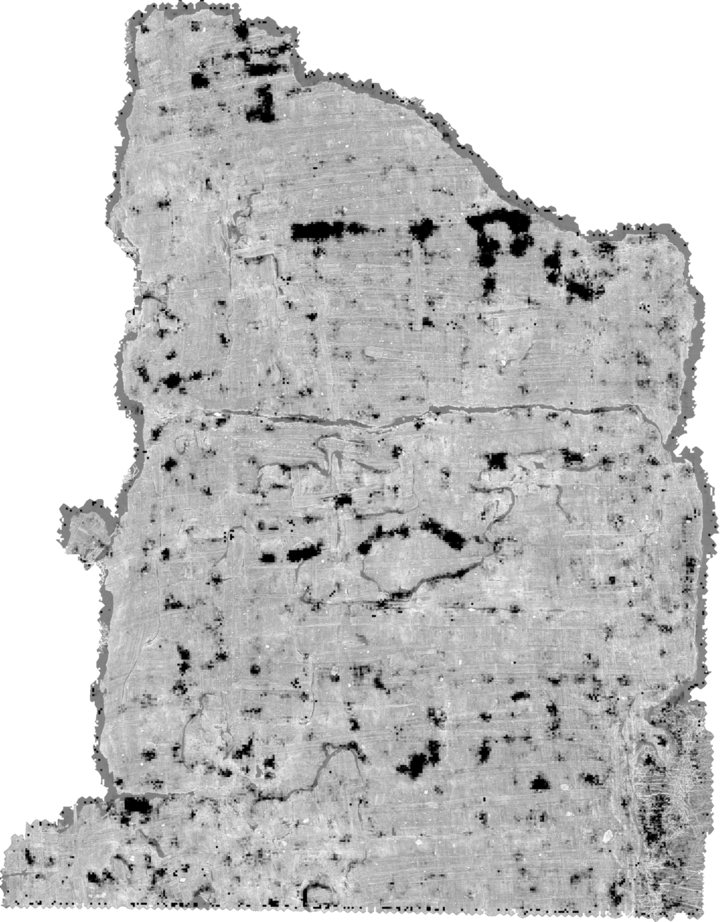}} &
        \addtolength{\tabcolsep}{-12pt}
            \raisebox{3\height}{
            \begin{tabular}{ll}
                1 & ].[ \\
                2 & ]\d{\textPi}\textRho.[ \\
                3 & ].\textAlpha.[ \\
                4 & ] \textit{ink traces} [ \\
            \end{tabular}
        } \\
        \addtolength{\tabcolsep}{12pt}
        \\

        \includegraphics[width=0.16\textwidth]{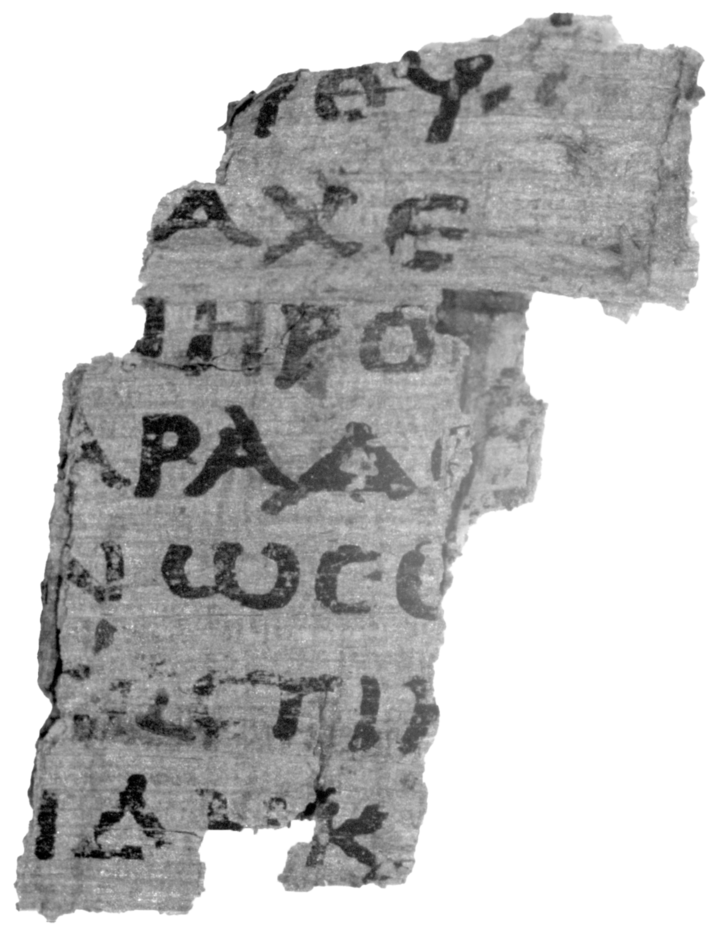} &
        \includegraphics[width=0.16\textwidth]{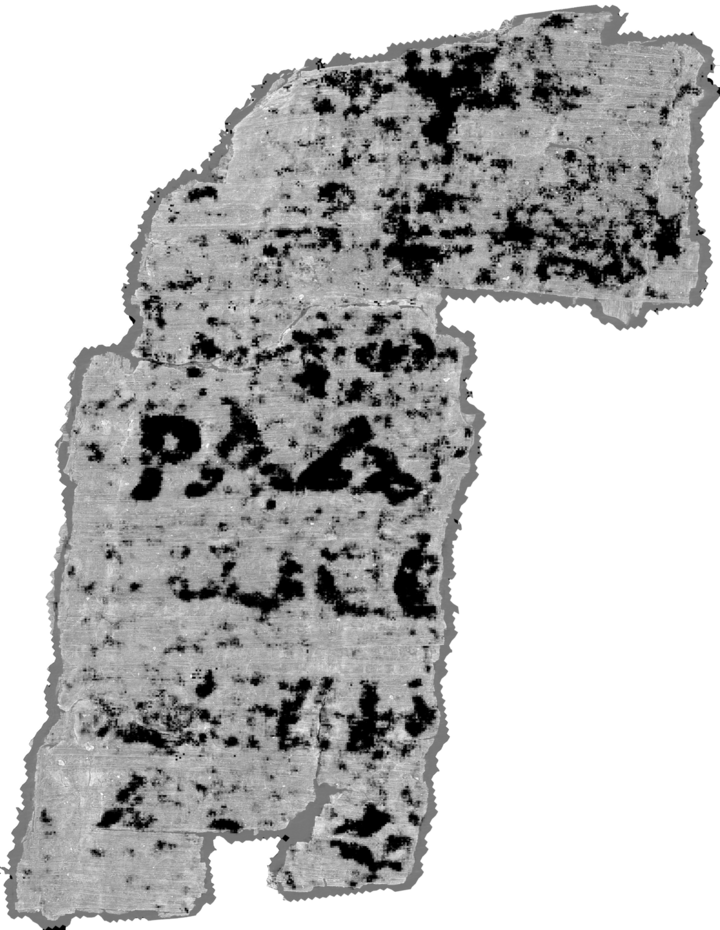} &
        \raisebox{0.2\height}{\includegraphics[width=0.12\textwidth]{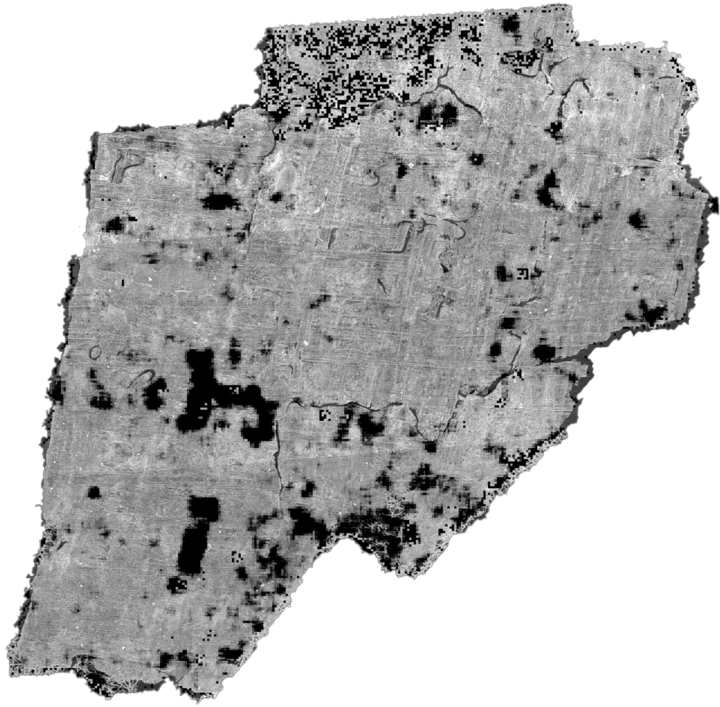}} &
        \addtolength{\tabcolsep}{-12pt}
            \raisebox{1.75\height}{
            \begin{tabular}{ll}
                1 & ]\textEta[ \\
                2 & ]\d{\textIota}[ \\
            \end{tabular}
        } \\
        \addtolength{\tabcolsep}{12pt}
        \\

        \includegraphics[width=0.16\textwidth]{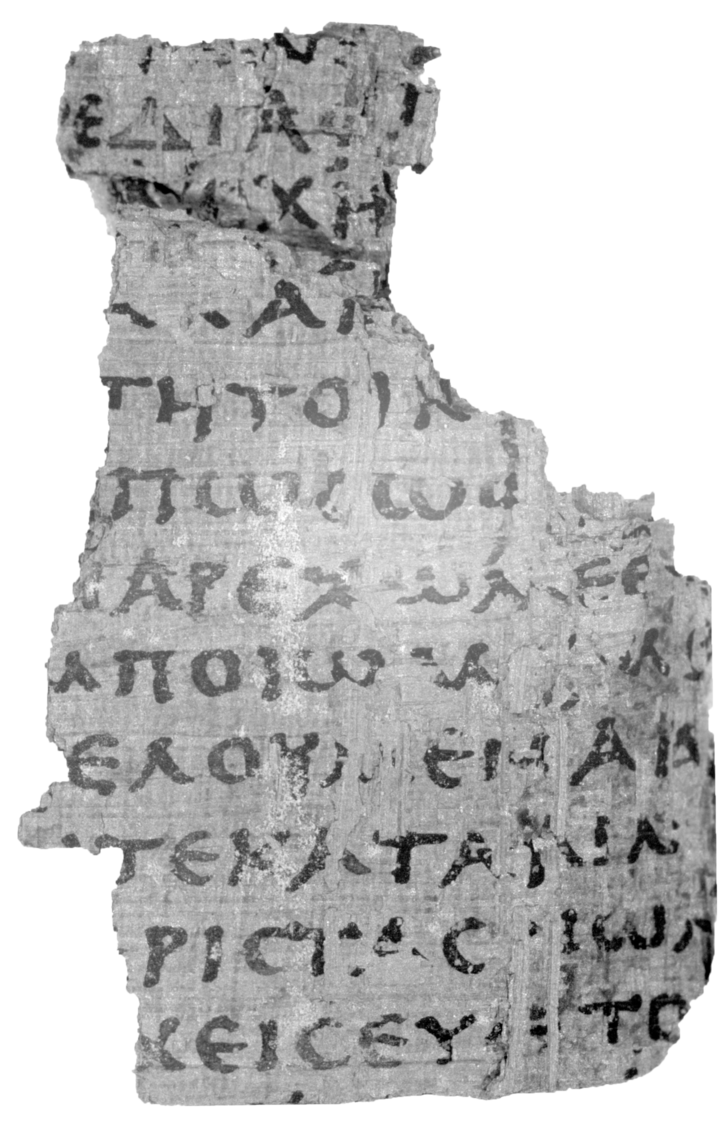} &
        \includegraphics[width=0.16\textwidth]{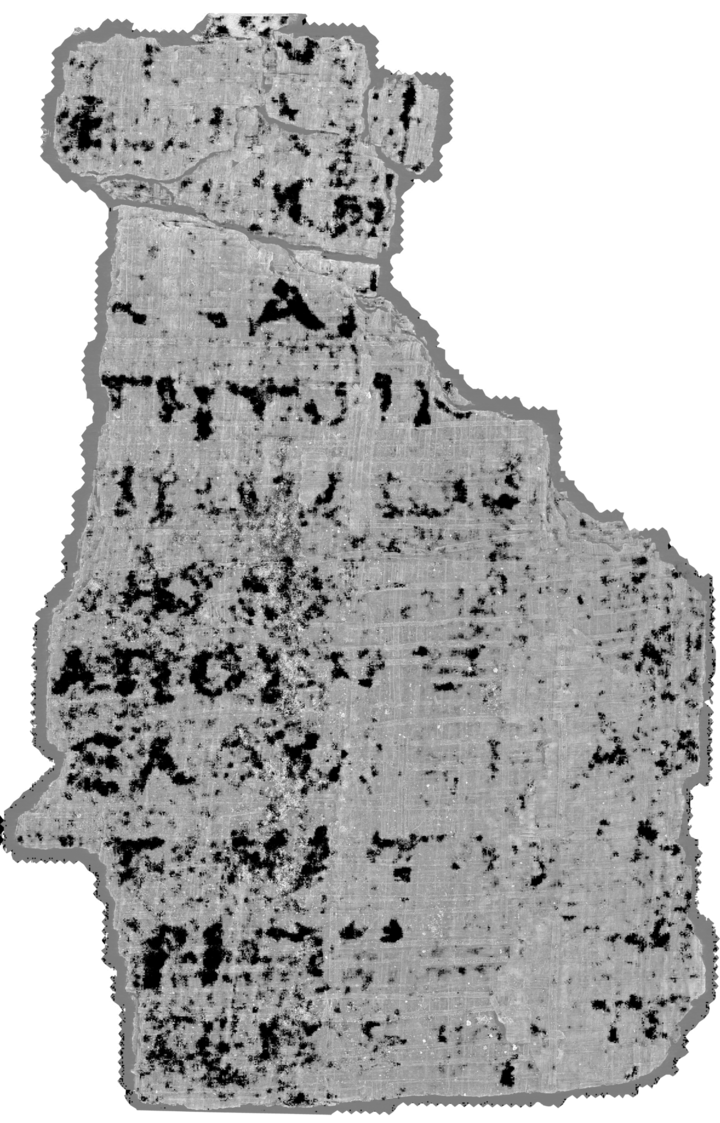} &
        \raisebox{0.6\height}{\includegraphics[width=0.12\textwidth]{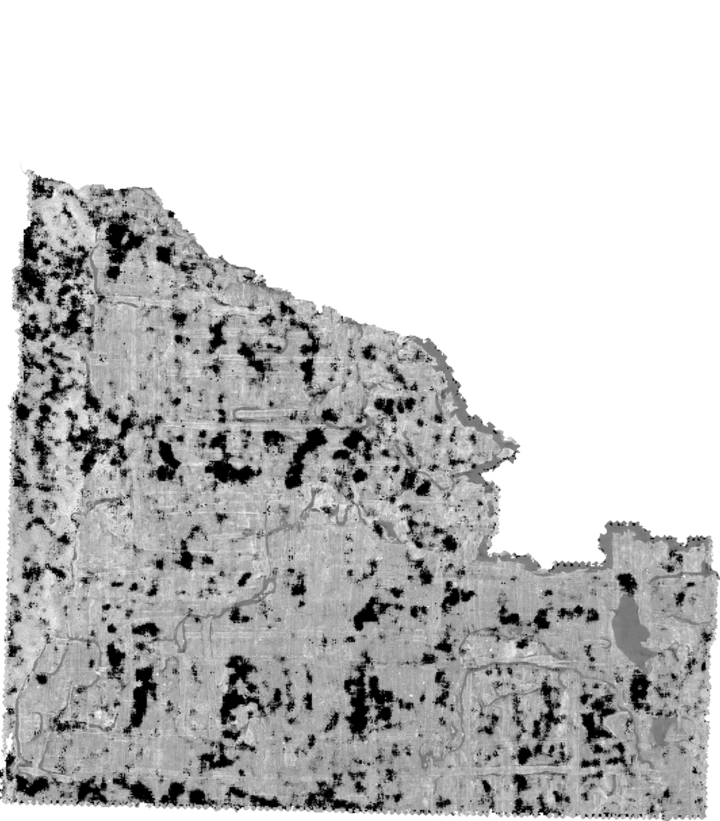}} &
        \addtolength{\tabcolsep}{-12pt}
            \raisebox{3.25\height}{
            \begin{tabular}{ll}
                1 & ] \textit{ink traces} [ \\
                2 & ]\textTau\d{\textEpsilon}\textUpsilon\d{C}[ \\
                3 & ] \textit{ink traces} [ \\
            \end{tabular}
        } \\
        \addtolength{\tabcolsep}{12pt}
        \\

        (a) Ground truth & 
        (b) Ours, surface & 
        (c) Ours, hidden &
        (d) (c), transcribed \\
    \end{tabular}
    \caption{Ink detection results for Herculaneum Fragments. (a) Ground truth infrared photographs of fragment surfaces. (b) Our method (Volume Cartographer + ink-ID) on fragment surfaces, generated purely from X-ray CT. Cross-validation used to prevent model memorization. (c) Our method on subsurface hidden layers, revealing text that has not been seen in nearly 2,000 years. (d) Greek transcriptions of (c). ] and [ indicate line beginning and end. Dot indicates indistinct ink traces, underdot indicates uncertain transcription.}
    \label{fig:fragments-results-overview}
\end{figure*}
\addtolength{\tabcolsep}{-12pt}

Using EduceLab-Scrolls, we present benchmark ink detection results on Herculaneum scroll fragments.
Figure \ref{fig:fragments-results-overview} shows the performance of our method, which is capable of learning the presence of carbon ink inside CT images of scroll fragments.
Figure \ref{fig:fragments-ground-truth} shows ground truth infrared images of the fragments, displaying the known true location of ink on their surfaces.
These fragments each come from a scroll that produced many such fragments, and scholars have identified the texts of these scrolls.
P.Herc.Paris.\ 1 fr.\ 34 (Papyrus Herculaneum Paris 1, fragment 34) and P.Herc.Paris.\ 1 fr.\ 39 come from a scroll about Hellenistic Dynastic history.
P.Herc.Paris.\ 2 fr.\ 47 and P.Herc.Paris.\ 2 fr.\ 143 are from a philosophical work authored by Philodemus, as are many others in the Herculaneum collection.
This book, titled \textit{On slander}, comes from his greater work \textit{On Vices and the opposite virtues}.
Our objective is to scan these fragments using X-ray CT and detect the ink inside those scans, starting by validating our results on the surface against the ground truth images.

Working solely from X-ray CT images, we apply the method to the exposed fragment surfaces, generating images showing the predicted location of carbon ink (Figure \ref{fig:fragments-ours-surface}).
Our method successfully reveals the correct location of much of the carbon ink on these surfaces, generating images with legible and accurate text.
Notably, though our method presently fails to detect some challenging ink spots, it does not hallucinate or generate substantial false positives that would mislead scholarly interpretation.

We next apply our method to hidden, subsurface layers of the scroll fragments for which there is no ground truth.
Shown in Figure \ref{fig:fragments-ours-hidden}, the ink detected by our method reveals Greek characters for these layers as well, recovering text that has not been seen in nearly 2,000 years.
Crucially, the scale, line separation, and script of the revealed characters are consistent with those observed on the fragment surfaces.

The results presented with this work represent the first time the hidden texts of Herculaneum have been recovered noninvasively.
The quantity and quality of the revealed text will both continue to increase as methods are refined and more data is acquired.
We believe this milestone proves the feasibility of noninvasive text recovery for the Herculaneum papyri, and marks the transition from proof of concept to scaling across the entire collection.

\subsection{Experimental setup}

The following high-level experimental setup was used for the results shown in Figure \ref{fig:fragments-results-overview}.
Details can be found in Appendix \ref{methods}.

The ground truth images in Figure \ref{fig:fragments-ground-truth} are 1000nm infrared photographs, acquired by Brigham Young University beginning in 1999 \cite{booras2001herculaneum,ware2000multispectral}.
This form of imaging and specific wavelength are chosen for the high contrast captured between ink and papyrus.

The images in Figure \ref{fig:fragments-ours-surface} are the result of a machine-learning based method for ink detection.
The model receives small 3D regions of X-ray CT as input, and learns to classify them as ``ink'' or ``not ink'' based on the presence of ink at their central voxel.
The ink labels used during supervised model training are created by labeling the infrared photographs of the fragment surfaces, creating binary ink label images.
An eight-fold cross-validation setup is used for Figure \ref{fig:fragments-ours-surface}, in which each of the four fragment surfaces is split into top and bottom halves, yielding eight total fragment regions.
Eight independent ink detection models are then trained, each one training on seven of the eight training regions but omitting one.
After training, the model generates an ink prediction image for the held-out region that it has not seen during training.

This ink prediction image is composited over the Volume Cartographer texture image: an intermediate product of the data processing pipeline which clearly reveals the papyrus fiber structure in X-ray but does not exhibit visible ink contrast.
The composite image, created simply by subtracting the ink prediction image from the texture image, creates an image of the prediction region that shows both the papyrus fiber structure of the surface, as well as the detected ink regions in black.
All eight of these images are then compiled together, yielding Figure \ref{fig:fragments-ours-surface}.
The result is a combined image showing the general ability of the learned models to detect carbon ink in X-ray CT.
Cross-validation is chosen in order to avoid model memorization, in which the model is asked to generate predictions for regions it has already seen during training.
Memorization would invalidate the ink detection results, as instead of learning to truly identify the presence of carbon ink, the model could simply remember what label was associated with a particular input.
Cross-validation instead ensures the model is being evaluated on its actual ability to generalize to unseen examples.
Of course, this is the ultimate objective of the model, as it will later be applied to hidden, subsurface layers for which there is no ground truth and it would be impossible to memorize.

Figure \ref{fig:fragments-ours-hidden} illustrates this mode, in which a trained model is applied to hidden layers extracted from the X-ray CT under the exposed surface.
In this case, there is no risk of memorization, so a model is trained on the entire training dataset, or all four fragment surfaces.
This one model is then used to generate the prediction images seen in Figure \ref{fig:fragments-ours-hidden}.

\subsection{Evaluation}

We verify the accuracy of our ink detection method against fragment surfaces with known ground truth.
In particular, we are interested in recall and false positives as they quantify subjective qualities of interest: how much of the true text can be recovered, and how much false text is generated in error?
While it is acceptable not to detect challenging ink regions, the model should not substantially ``hallucinate''; that is, it should not generate misleading images of characters that are not actually present.

\subsubsection{Visual}

Visual confirmation allows even a non-expert to quickly verify the generated images against the ground truth, checking for rough accuracy, recall, and noise in the generated images.
This validation is used extensively in the development cycle as an early indicator of model performance.
Visual inspection allows one to evaluate multiple subjective qualities that combine to create ``readability,'' a measure that in our experience does not always correlate precisely with quantitative metrics.
A brief visual comparison of Figure \ref{fig:fragments-ours-surface} against the ground truth in Figure \ref{fig:fragments-ground-truth} suggests that the method is correctly detecting the locations of text lines, nearing $\approx50\%$ recall of individual characters, and without notable false positives (characters that do not exist in the ground truth).

\subsubsection{Quantitative pixel-based metrics}

We also report the quantitative performance of our method using image comparison metrics.
As this is a new task and dataset, these are presented primarily as a benchmark against which future improvements can be evaluated.

\begin{figure}
    \centering

    \begin{subfigure}{0\textwidth}
        \refstepcounter{subfigure}\label{fig:binary-comparison-a}
    \end{subfigure}
    \begin{subfigure}{0\textwidth}
        \refstepcounter{subfigure}\label{fig:binary-comparison-b}
    \end{subfigure}

    \begin{tabular}{cc}
        \includegraphics[width=0.4\linewidth]{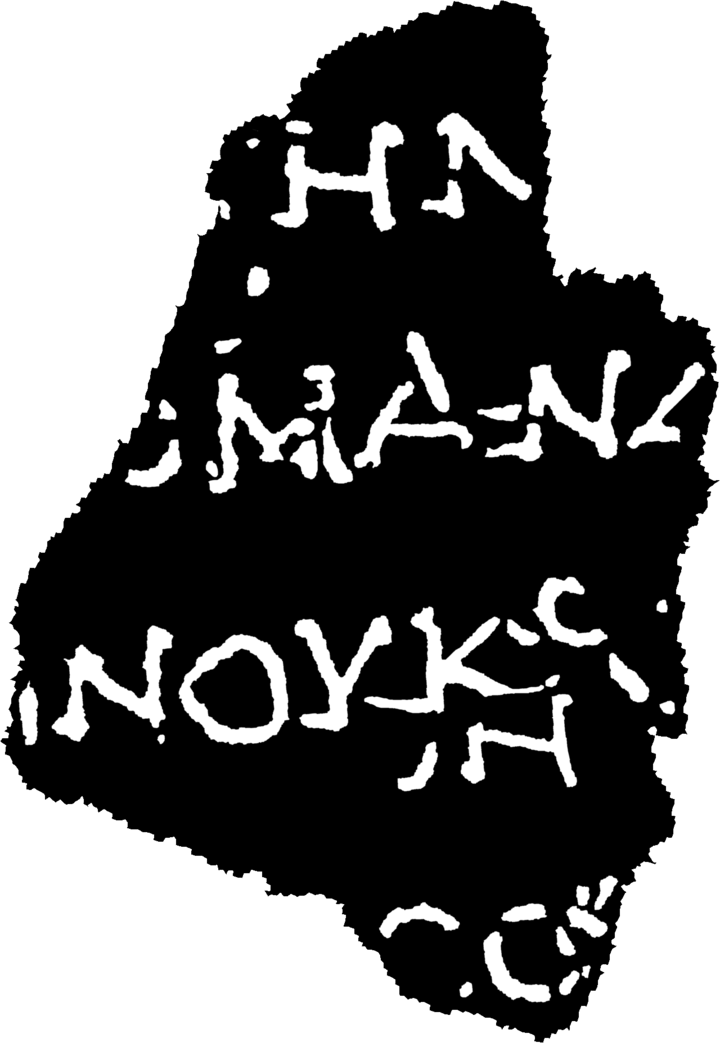} &
        \includegraphics[width=0.4\linewidth]{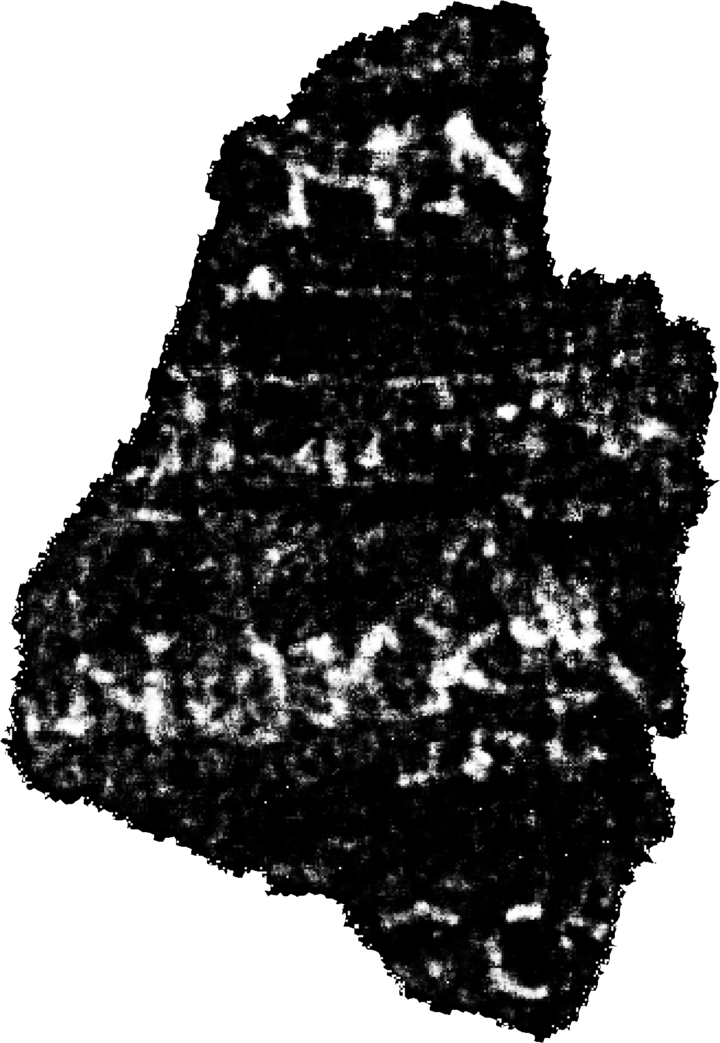} \\

        (a) Ground truth &
        (b) Ours \\
    \end{tabular}
    \caption{Ground truth and our output for the binary ink classification task on P.Herc.Paris.\ 1 fr.\ 34.}
    \label{fig:binary-comparison}
\end{figure}

Ink identification is performed as a binary classification task, with predictions generated for each output pixel individually.
Figure \ref{fig:binary-comparison} visualizes this task, showing the ground truth binary label mask (\ref{fig:binary-comparison-a}) and our generated prediction image (\ref{fig:binary-comparison-b}).
Once pixel predictions are aggregated at the image level, they can be evaluated against the ground truth using common metrics from semantic segmentation, another common computer vision task.

The widely used binary cross entropy and Dice \cite{dice1945measures} (also known as F1) metrics are reported.
Precision, recall and false positive rate (FPR) are also included for their aforementioned ties to subjective qualities of interest.
Additionally reported is F0.5, a measure similar to Dice/F1 that weights precision higher than recall in order to reduce false positives and encourage sharper characters:
\begin{equation}
    \frac{(1+\beta^2)pr}{\beta^2 p + r},\quad \beta=0.5
\end{equation}
where $p$ is precision and $r$ is recall.

\begin{table}
    \centering
    \begin{tabular}{lll}
        \toprule
        Metric & \textmu{} & $\sigma$ \\
        \midrule
        Binary cross entropy & 0.44 & \num{5.6e-3} \\
        Precision & 0.58 & \num{4.2e-2} \\
        Recall & 0.41 & \num{3.0e-2} \\
        FPR & 0.051 & \num{9.9e-3} \\
        Dice/F1 & 0.48 & \num{1.5e-2} \\
        F0.5 & 0.53 & \num{2.4e-2} \\
        \bottomrule
    \end{tabular}
    \caption{Aggregate validation results for the pixel-wise binary ink detection task, across four Herculaneum fragments, following 8-fold cross-validation. Mean (\textmu{}) and standard deviation ($\sigma$) reported for each, sampled from the second half of 620,000 total training batches.}
    \label{table:quantitative-results}
\end{table}

Table \ref{table:quantitative-results} shows the metrics associated with the results visualized in Figure \ref{fig:fragments-results-overview}.
The results are aggregated across the 8-fold cross validation, meaning the 8 individual predictions are compiled and then metrics are computed.
As output predictions fluctuate throughout training, the mean \textmu{} and standard deviation $\sigma$ are reported, taken from the second half of 620,000 training batches.
The second half is chosen as the model converges before this point.

The quantitative metrics agree with the visual results, with recall higher than the false positive rate by a factor of 8.
Both visual and quantitative results suggest the model is capable of accurate ink detection without significant hallucination.

\subsubsection{Papyrological character-based metrics}

Quantitative image metrics are helpful for method development, as they can be automated and provide quick feedback.
However, they do not exactly capture the ultimate objective, which is the readability of generated text to the trained papyrologists who study the Herculaneum papyri.
Again we measure recall and false positives, this time with respect to individual characters as identified by papyrologists.

We provided our generated images of the fragments (Figure \ref{fig:fragments-ours-surface}) to a Herculaneum papyrologist, who was previously unfamiliar with these fragments.
They transcribed each fragment based solely on the provided images, which are generated purely from X-ray CT.
In all cases, the scholar was able to identify the correct number of lines of the fragment surface, even correctly identifying in P.Herc.Paris.\ 1 fr.\ 34 where the surface is broken and marks a transition between two separate writing layers.

\begin{figure}
    \centering
    \begin{tabular}{ll@{\hspace{3em}}ll}
        & layer a & & layer a \\
        \\
        1 & ].\textEta\textNu[ & 1 & ].\textEta\textNu[ \\
        2 & ]\d{\textOmikron}\textMu\textAlpha\textNu\textDelta[ & 2 & ].....[ \\
        3 & ].\textNu\textOmikron\textUpsilon\textKappa.[ & 3 & ].\textNu\textOmikron\textUpsilon\textKappa.[ \\
        \\
        & layer b & & layer b \\
        \\
        1 & ]\d{\textOmikron}\textNu[ & 1 & ]..[ \\
        2 & ]\d{\textEpsilon}\d{C}.[ & 2 & ].\d{C}.[ \\
        \\
        & (a) Ground truth & & (b) Ours \\
    \end{tabular}
    \caption{Greek transcriptions for P.Herc.Paris.\ 1 fr.\ 34 from a trained papyrologist, of both the ground truth image and our generated image. ] and [ indicate line beginning and end. A dot indicates indistinct ink traces and an underdot indicates an uncertain transcription.}
    \label{fig:papyrological-review}
\end{figure}

Figure \ref{fig:papyrological-review} shows the P.Herc.Paris.\ 1 fr.\ 34 transcriptions from the ground truth and generated images, respectively.
Of the 15 characters identified in the ground truth, 7 are recovered accurately using our method.
No characters are misidentified, or identified where none exist.

\begin{table}
    \centering
    \begin{tabular}{llll}
        \toprule
        Fragment & Characters & Recall & FPR \\
        \midrule
        1 & 26 & 0.23 & 0.27 \\
        2 & 59 & 0.41 & 0.22 \\
        3 & 15 & 0.47 & 0.00 \\
        4 & 25 & 0.64 & 0.12 \\
        Cumulative & 125 & 0.42 & 0.18 \\
        \bottomrule
    \end{tabular}
    \caption{Character recall and false positive rate (FPR) across four Herculaneum fragments, comparing human transcriptions from our generated images against human transcriptions from ground truth. The number of characters in the ground truth transcription is also shown.}
    \label{table:papyrological-review}
\end{table}

Table \ref{table:papyrological-review} summarizes these results across four Herculaneum fragments.
In all fragments but P.Herc.Paris.\ 2 fr.\ 47, recall exceeds the false positive rate by at least a factor of 2.
The results are computed using a ``strict'' scoring, in which even those characters identified as uncertain by the papyrologist are scored.
Excluding these characters would further improve the results.
Further, characters that are mistakenly identified are typically the result of partial ink recovery, showing only some of the ink strokes making up a complete character.
Importantly, the misidentified characters are not invented ``whole cloth.''
They are at least identified in the correct location, and could be accurately identified with marginal improvements in ink detection.

These results are a tremendous step forward from the previously nonexistent baseline.
This method already performs well enough to generate valuable text for Herculaneum papyrologists, who are used to working with fragmentary writing even in the best cases.
That said, there is clear room for improved performance, which we anticipate will follow this work.

\section{Related work}

EduceLab-Scrolls builds on a wide-ranging body of work across heritage science, machine learning, and applied computational methods.

\subsection{Virtual unwrapping}

Virtual unwrapping \cite{seales2003digital, seales2017invisible} has now been established as a prominent intersection of heritage science and computational methods, and there are a growing number of successful applications.
Virtual unwrapping involves noninvasive volumetric imaging followed by a multi-step computational pipeline that traces the 3D geometry of the surfaces (segmentation), filters the volumetric image to make writing appear (texturing), and maps the result to a 2D image that is easily read (flattening).
The details of these steps and their ordering can vary to adapt to different challenges.

Multiple works have applied virtual unwrapping as a functional proof of concept on lab-made proxy manuscripts, illustrating the potential across a range of imaging methods, writing substrates, inks, and manuscript forms such as scrolls, codices, and folded sheets.
Early work focused on proxy materials with exaggerated features, validating the computational concept using much lower imaging resolution than is now available \cite{seales2004digital, lin2005opaque, lin2007physically}.
Since then, virtual unwrapping has worked for bamboo scrolls \cite{stromer2018non, stromer2019virtual}, metallic inks on parchment \cite{stromer2017browsing, stromer2018browsing}, rolled and folded papyri \cite{allegra2016x, baum2017revealing}, and even specifically carbon ink on papyrus in the right conditions \cite{parker2019inkid, tserevelakis2018uncovering}.

Building on the learnings from lab-made proxy manuscripts, other works have gone on to recover text from genuine heritage manuscripts.
These also span a wide range of materials, including metallic inks on parchment, paper, and papyri \cite{baumann2008use, dilley2022m910, samko2014virtual, liu2018robust, rosin2018virtual, seales2016engedi, mahnke2020virtual, dambrogio2021unlocking, albertin2022xray}, etched metal scrolls \cite{barfod2015revealing, baum2021revisiting}, and lead amulets \cite{wilsterhansen2021virtual}.
Some works \cite{stabile2021computational} focus specifically on segmentation, leaving ink detection as future work.
Ultimately, the goal of developing virtual unwrapping methods is to recover substantial texts leading to new material for scholarly publication, an objective achieved with the En-Gedi scroll \cite{seales2016engedi, segal2016engedi}.

The above works vary significantly in the segmentation approach and in the imaging method.
The segmentation approach varies in order to handle different materials and their respective geometries.
Imaging techniques are typically varied in order to generate high contrast between the ink and substrate, if possible.
Relying on this strong contrast, texturing methods are consistently simple local filters that directly extract the image intensities in a local neighborhood.
Recently, this has shifted with the introduction of machine learning-based texturing methods that detect subtle signals \cite{parker2019inkid} and learn to map the volumetric input to other image domains \cite{parsons2020expressive}.

\subsection{Noninvasive study of Herculaneum papyri}

For fragments with exposed writing, strong visual contrast is possible through the use of spectral imaging in the infrared bands \cite{booras2001herculaneum,ware2000multispectral}.
For a rare Herculaneum fragment with text on the verso, or the back side of the writing surface, the use of shortwave-infrared (1000-2500 nm) imaging has even enabled the recovery of this writing \cite{tournie2019ancient}.
While exciting for the few exposed fragments with verso text, this method has limited penetration into the papyrus and would not extend to the hidden layers within a rolled scroll.

Considerable work has investigated the chemical composition of the ink of the Herculaneum papyri fragments, with the goal of informing future imaging methods that would be able to capture clear ink contrast from within an intact scroll.
There have been mixed results.
Early investigations \cite{seales2013virtual} used at least nine different imaging technologies on three Herculaneum fragments, looking for signals that would differentiate ink from papyrus.
Scanning electron microscopy and X-ray fluorescence found calcium in the ink but not papyrus, and particle-induced X-ray emission suggested the trace presence of lead and strontium in the ink only.

Other work with different Herculaneum fragments has discovered a stronger presence of lead in the ink, leading to clear imaging contrast using X-ray fluorescence \cite{brun2016revealing, tack2016tracking}.
The reason for the lead presence is not yet known, but could arise from contamination or from deliberate use as a pigment, binding, or drying agent.
It is also not known why the ink of some fragments exhibits a strong lead profile while others do not, though this is not too surprising, considering the Herculaneum papyri were authored by different scribes, using homemade inks, over a period of three centuries.
A more recent and thorough X-ray fluorescence study of 38 Herculaneum scroll fragments validated the other studies, finding multiple elements that sometimes correlated strongly with ink: phosphorus, 5 fragments; iron, 3; copper, 3; and lead, 2 \cite{bonnerot2020xrf}.
Carbon-based inks not specific to the Herculaneum papyri have also been studied thoroughly, with similar findings and implications \cite{gibson2018assessment, christiansen2020insights, autran2021revealing}.

Imaging of the intact Herculaneum scrolls has also matured, despite not yet achieving clear ink contrast.
The first X-ray CT images of intact Herculaneum scrolls revealed the internal structure but no ink contrast \cite{seales2009lire, seales2011analysis}.
Phase contrast X-ray CT was also proposed \cite{seales2013virtual} and then conducted \cite{mocella2015revealing, bukreeva2016virtual, bukreeva2017investigating} as a potential technique to achieve ink contrast inside a rolled scroll.
Despite early claims of textual discovery, this technique has not led to further discoveries or ongoing scholarly work.
The most recent imaging contains implicit phase shift data, but did not prioritize the amplification of this shift, more resembling standard X-ray micro-CT.
These images and their processing, released in EduceLab-Scrolls, instead represent a focus on the salient cues we so far know are crucial to ink detection: the highest achievable resolution, precise segmentation, and accurate labeling.
These factors combine to create a dataset in which machine learning-based methods can detect the ink presence, even without strong visual contrast.

We view the research and development of methods for improved imaging contrast as highly complementary to the work we present here.
Though machine learning-based methods are now capable of detecting Herculaneum ink in X-ray CT images, there remains room for improvement.
We anticipate this will happen in part due solely to improved software pipelines using existing data, but any improvement in imaging contrast would greatly boost the ultimate accuracy of these pipelines.
The additional information captured by phase contrast or other techniques can ultimately only help, as trained models can select the most useful features from the input images.

\subsection{Volumetric datasets}

Volumetric datasets are common in the medical domain, but are of lower resolution due to the radiation dosage and motion constraints of medical imaging.
Methods developed for volumetric medical images nonetheless do have significant overlap with those of virtual unwrapping, for example in surface segmentation \cite{li2005optimal}.

One of the most similar domains in terms of volumetric datasets and tools is connectomics, or the study of connections in the nervous system.
Like the challenge of the Herculaneum papyri, tracing the connections of neurons requires an unusual combination of volumetric image resolution and dimensions.
The resulting large datasets have generated a growing ecosystem of file formats \cite{jeremy_maitin_shepard_2021_5573294}, web viewers \cite{jeremy_maitin_shepard_2021_5573294}, cloud volume processing frameworks \cite{william_silversmith_2021_5671443}, and more.
Today, both connectomics and virtual unwrapping use specialized toolsets developed for particular domains or even specific problems within a domain.
We anticipate that with recent successes in both fields, the tools will generalize and become more powerful.

Even the most similar methods in these other domains still rely on expert labeling for supervised tasks, in which a human directly labels the volumetric image.
We believe our geometric framework is the first of its kind, through which labels can be injected into the 3D volume from a 2D image acquired in another modality.

\section{The EduceLab-Scrolls dataset}

\subsection{Design}

EduceLab-Scrolls is designed to be a comprehensive dataset, with sufficient training and inference data to recover real texts from the hidden papyrus layers of the Herculaneum papyri collection.
``Sufficient'' here concerns both quality and quantity.
Further, much work is put into the geometry of the resulting images, such that a training dataset can be constructed.

\subsubsection{Quality}

At a high level, sufficient quality means that the X-ray CT volumes capture the ink presence in a way that is detectable using learned methods.
In practice, the two primary factors are the scan image resolution and the energy of the incident X-ray beam.
For many virtual unwrapping projects, imaging voxel sizes in the range of 20-100\textmu{}m are sufficient for text extraction.
This is due to the strong contrast achieved in most cases, in which the dense, often metallic components of the ink create a bright response in X-ray CT.
Metallic inks are capable of creating a brightness response in X-ray even when the image voxel size is itself larger than the thickness of the ink layer.
With current imaging methods, this form of contrast has not been achieved with the Herculaneum papyri.
Instead, a much more subtle signal is captured, likely relating not only to image intensity but to morphological (texture, or shape) differences between ink and blank papyrus.
As prior work has inferred the Herculaneum ink thickness to be on the order of 5-10\textmu{}m \cite{parker2019inkid}, CT image voxel sizes must approach this threshold in order to meaningfully resolve the detail needed for ink detection.
This threshold is frequently on the edge of feasibility using existing imaging techniques.
The CT images in EduceLab-Scrolls therefore consistently represent the highest achievable resolution (smallest voxel size) for a given scan session.
Available imaging technology at the time, physical object dimensions, practical time constraints during scanning, and other factors result in CT scans that range from 2.23-23.9\textmu{}m.
Those datasets that we find most promising are more tightly clustered between 3.24-7.91\textmu{}m.

The relationship between incident X-ray energy and achievable carbon ink detection is less understood, though other work \cite{stromer2018browsing} as well as anecdotal evidence from our own experience suggest that lower-energy CT scans better capture the subtle contrast of low-Z elements such as carbon.
Therefore, the scans in EduceLab-Scrolls largely utilize the lowest achievable incident X-ray energy that enables detailed reconstructions, as low as 22kV.
The differing units in this range hint at another difference between X-ray beams: benchtop X-ray sources use an X-ray lamp set to a particular kV in order to generate a polychromatic X-ray beam with a wide distribution, while synchrotron sources are capable of monochromatic X-ray beams at a specific keV.
It is not yet understood which form of X-ray source best captures the ink signal, though we have observed that both are sufficient for ink detection.

\subsubsection{Quantity}

The ink of the Herculaneum papyri is known to have variable thickness and composition, so there should be adequate training data to capture this distribution such that models can generalize to the ink of unseen hidden layers.
EduceLab-Scrolls contains all labeled training data generated to date, spanning 16 scans of 6 fragments from 4 scrolls.
The training fragments themselves are each comprised of multiple papyrus layers, so in addition to exposed surfaces they contain hidden papyrus layers which are candidates for inference and therefore the discovery of previously hidden texts.
The dataset also contains all releasable scans of intact Herculaneum scrolls to date: 9 CT scans spanning 4 scrolls.
The quantity of training and inference data is sufficient for the textual results presented in this work, and we anticipate EduceLab-Scrolls in its current form contains much more text that will be recovered from the existing dataset.
We further anticipate that these results will only improve as more training and inference data are eventually acquired and processed.

\subsubsection{Geometry}

\begin{figure*}
    \centering
    \captionsetup{justification=centering}

    \begin{subfigure}{0\textwidth}
        \refstepcounter{subfigure}\label{fig:overview-a}
    \end{subfigure}
    \begin{subfigure}{0\textwidth}
        \refstepcounter{subfigure}\label{fig:overview-b}
    \end{subfigure}
    \begin{subfigure}{0\textwidth}
        \refstepcounter{subfigure}\label{fig:overview-c}
    \end{subfigure}
    \begin{subfigure}{0\textwidth}
        \refstepcounter{subfigure}\label{fig:overview-d}
    \end{subfigure}
    \begin{subfigure}{0\textwidth}
        \refstepcounter{subfigure}\label{fig:overview-e}
    \end{subfigure}
    \begin{subfigure}{0\textwidth}
        \refstepcounter{subfigure}\label{fig:overview-f}
    \end{subfigure}
    \begin{subfigure}{0\textwidth}
        \refstepcounter{subfigure}\label{fig:overview-g}
    \end{subfigure}

    \includegraphics[width=\textwidth]{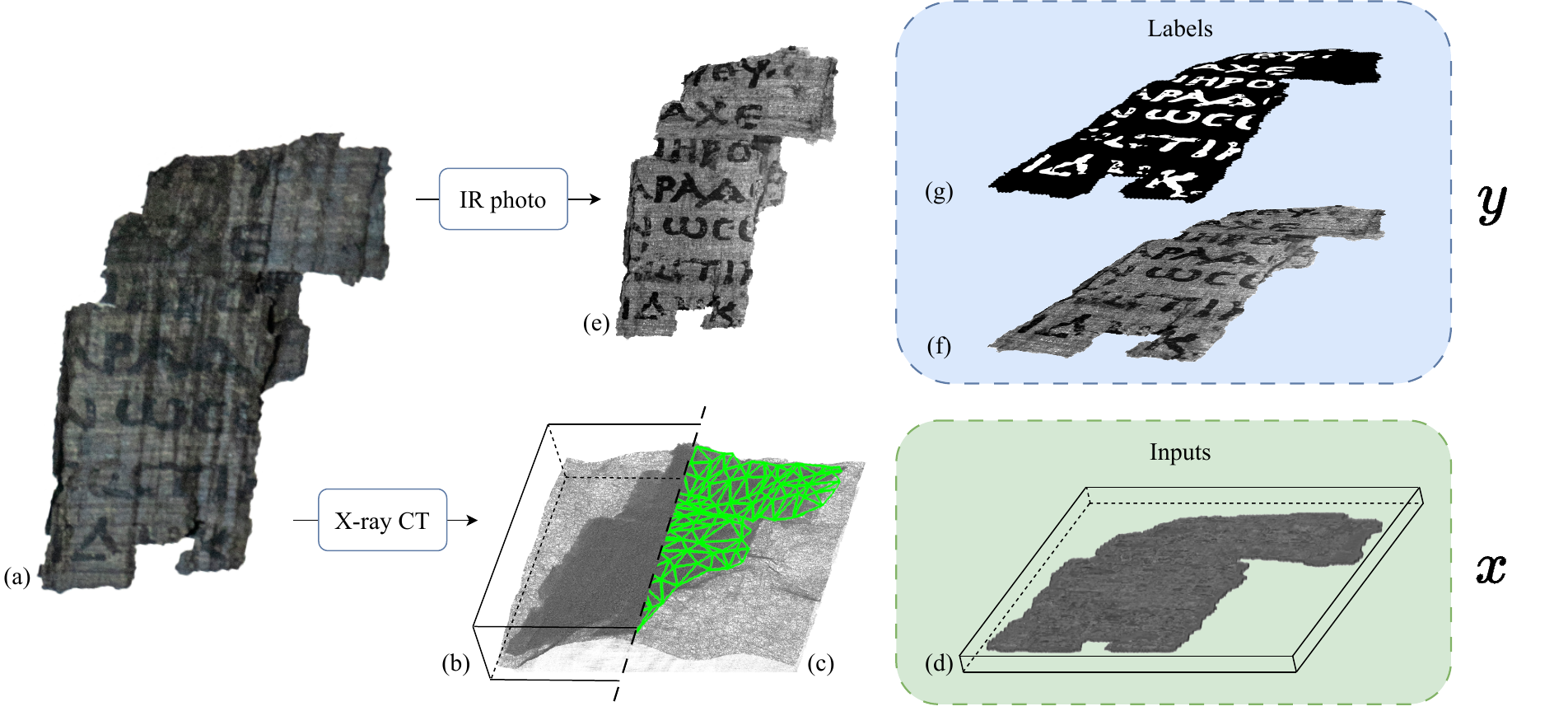}
    \caption{EduceLab-Scrolls dataset geometry for an example fragment. (a) Scroll fragment, RGB photograph. (b) Volumetric X-ray CT image. (c) 3D surface segmentation. (d) Flattened ``surface volume'' sampled about the segmented surface mesh. (e) Infrared photograph. (f) Infrared photograph aligned to surface volume. (g) Aligned binary ink labels.}
    \label{fig:method-overview}
\end{figure*}

The EduceLab-Scrolls dataset is designed to enable supervised machine learning across 3D volumetric input images, where labels are available only from 2D surface images.
The training dataset is constructed from those objects for which labeling is possible, i.e. the detached scroll fragments with exposed surface text.
For these objects, labels are generated and aligned to volumetric inputs in a process outlined in Figure \ref{fig:method-overview}.
To reveal the ink of hidden, subsurface layers, volumetric inputs without labels are created using steps (a-d) of Figure \ref{fig:method-overview}, and a model trained on the training dataset performs inference on these volumes.

The label alignment and generation process relies on finding a shared image space, to which both the 3D volumetric images and the 2D surface images can be aligned.
The chosen space is the 2D image space created when the segmented surface is flattened, a transition shown from Figure \ref{fig:overview-c} to \ref{fig:overview-d}.
The ``surface volume'' shown in Figure \ref{fig:overview-d} is then sampled from the original volume into this flattened space.
The resulting image is volumetric and 3D, but can also be considered as a 2D image with multiple depth channels.
An infrared surface photograph (Figure \ref{fig:overview-e}) is then aligned (Figure \ref{fig:overview-f}) to this surface volume, and a binary ink label image is created (Figure \ref{fig:overview-g}).
The surface volume and ink label now serve as an (input, label) image pair for supervised learning.
This process is discussed in more detail in Appendix \ref{methods} and is presented here only to illustrate the high level dataset design.

\subsection{Dataset overview}

Table \ref{table:overview-fragments} contains an overview of the EduceLab-Scrolls dataset.
The field of view, voxel size, incident energy, and disk size are listed for each CT scan included.
Most fragments and scrolls are scanned at multiple energies, an exciting and under-explored opportunity that we believe will lead to improved ink detection once leveraged.
Each fragment surface in the dataset also includes the full set of intermediate files from the data processing, representing the steps of the segmentation and labeling phases.
We believe this is the first dataset of its kind and the largest dataset released in the heritage domain.
Though the total disk size of the dataset exceeds 11TB, the ink detection results in this paper only require the segmented surface volumes, totaling 55.8GB.

\begin{table*}
    \centering
    \captionsetup{justification=centering}

    \begin{tabular}{rrrrrrr}
        \toprule
        & FOV & Scan year & Voxel size (\textmu{}m) & Energy & Size (GB) & Surface volume (GB) \\
        \midrule
        \\
        \multicolumn{1}{l}{Fragments - labeled surfaces} \\
        \midrule
        P.Herc.Paris.\ 1 fr.\ 34 & Full & 2019 & 3.2 & 54keV & 113 & 4.9 \\
        P.Herc.Paris.\ 1 fr.\ 34 & Full & 2019 & 3.2 & 88keV & 125 & TBD \\
        P.Herc.Paris.\ 1 fr.\ 39 & Full & 2019 & 3.2 & 54keV & 198 & 7.3 \\
        P.Herc.Paris.\ 1 fr.\ 39 & Full & 2019 & 3.2 & 88keV & 202 & TBD \\
        P.Herc.Paris.\ 2 fr.\ 47 & Full & 2019 & 3.2 & 54keV & 136 & 6.3 \\
        P.Herc.Paris.\ 2 fr.\ 47 & Full & 2019 & 3.2 & 88keV & 159 & TBD \\
        P.Herc.Paris.\ 2 fr.\ 143 & Full & 2019 & 3.2 & 54keV & 602 & 18 \\
        P.Herc.Paris.\ 2 fr.\ 143 & Full & 2019 & 3.2 & 88keV & 560 & TBD \\
        P.Herc.Paris.\ Objet 59 & Full & 2019 & 3.4 & 22kV & 145 & TBD \\
        P.Herc.Paris.\ Objet 59 & Full & 2019 & 3.4 & 29kV & 143 & TBD \\
        P.Herc.Paris.\ Objet 59 & Full & 2019 & 3.4 & 40kV & 145 & TBD \\
        P.Herc.Paris.\ Objet 59 & Full & 2019 & 3.4 & 59kV & 136 & TBD \\
        P.Herc.\ s.n.\ (Cass.\ 110) & Center & 2018 & 2.2 & 44kV & 73 & TBD \\
        P.Herc.\ s.n.\ (Cass.\ 110) & Center & 2018 & 11.9 & 60kV & 5 & TBD \\
        P.Herc.\ s.n.\ (Cass.\ 110) & Full & 2018 & 5.1 & 44kV & 171 & TBD \\
        P.Herc.\ s.n.\ (Cass.\ 110) & Full & 2018 & 5.1 & 49kV & 213 & TBD \\
        \\
        \multicolumn{1}{l}{Fragments - hidden layers} \\
        \midrule
        P.Herc.Paris.\ 1 fr.\ 39 & -- & 2019 & 3.2 & 54keV & -- & 5.1 \\
        P.Herc.Paris.\ 2 fr.\ 47 & -- & 2019 & 3.2 & 54keV & -- & 3.2 \\
        P.Herc.Paris.\ 2 fr.\ 143 & -- & 2019 & 3.2 & 54keV & -- & 11 \\
        \\
        \multicolumn{1}{l}{Intact scrolls} \\
        \midrule
        P.Herc.Paris.\ 3 & Full & 2009 & 27.1 & 70 & 277 & -- \\
        P.Herc.Paris.\ 3 & Half 1 & 2019 & 7.9 & 54keV & 3200 & -- \\
        P.Herc.Paris.\ 3 & Half 2 & 2019 & 7.9 & 54keV & 1200 & -- \\
        P.Herc.Paris.\ 3 & Center & 2019 & 7.9 & 88keV & 284 & -- \\
        P.Herc.Paris.\ 4 & Full & 2009 & 23.9 & 70kV & 107 & -- \\
        P.Herc.Paris.\ 4 & Half 1 & 2019 & 7.9 & 54keV & 1700 & -- \\
        P.Herc.Paris.\ 4 & Half 2 & 2019 & 7.9 & 54keV & 1300 & -- \\
        P.Herc.\ 803 & Most & 2019 & 23.9 & 70kV & 151 & -- \\
        P.Herc.\ 804 & Full & 2019 & 23.9 & 70kV & 206 & -- \\
        \bottomrule
        & & & & & 11.5TB & 55.8GB \\
    \end{tabular}
    \caption{
        EduceLab-Scrolls dataset overview.
        FOV: Field of view.
        keV indicates monochromatic synchrotron X-ray source, kV indicates polychromatic benchtop source.
        Some names are temporarily obfuscated.
    }
    \label{table:overview-fragments}
\end{table*}

\subsection{Availability}

EduceLab-Scrolls will be released under the Creative Commons Attribution-NonCommercial 4.0 Public License.
A subset of EduceLab-Scrolls is already available under a different license through Vesuvius Challenge, a machine learning and computer vision competition to build on this work and recover more text from the Herculaneum papyri.
For more information, visit \href{https://scrollprize.org}{scrollprize.org}.

\section{Conclusion}

We have presented EduceLab-Scrolls, a first of its kind dataset of Herculaneum scroll images that combines volumetric X-ray CT with aligned 2D labels.
Together with our software pipeline, which combines machine learning with a novel geometric framework, the combined data and method are capable of detecting the carbon ink inside Herculaneum scrolls.
Using scroll fragments with known ground truth, we validate the method using multiple metrics, including an assessment by Herculaneum scholars which confirms the accuracy of our generated images.
We believe this approach represents the definitive pathway to reading the intact Herculaneum scrolls using current imaging technology and computational methods.

We imagine broad future work in this research area.
Many exciting tools and advancements across virtual unwrapping and volumetric imaging, including this one, have been developed for specific applied cases.
With growing successes, there is now a significant opportunity to combine some of these tools into more generalizable and flexible frameworks, making these methods accessible to a larger pool of users.

With the discovery of hidden Herculaneum texts becoming a reality, it will also be increasingly critical to develop methods that are verifiable and reproducible \cite{brusuelas2021scholarly, chapman2021mets}.
Via fragments with verifiable ground truth, open data release, and open source code, we aim to achieve this with EduceLab-Scrolls and our method.
We further aspire to ease the reproducibility of these methods as we continue to develop them.

Of course, there remains a clear gap between the results presented here and the ultimate goal of reading intact Herculaneum scrolls in full.
We are optimistic we will together close this gap, as there are many opportunities for improvements in the existing pipeline.
Improved ink detection is a product of imaging, segmentation, alignment, labeling, and classification tasks.
EduceLab-Scrolls and our method provide a baseline for all of these, but each of them can be independently refined to improve the final result.
These improvements can also now be quantified, thanks to the introduction of labeled ground truth images from fragment surfaces.
In our experience, small improvements to the various stages of the pipeline combine to yield major improvements in text readability.
With time, we believe these cumulative improvements will not only recover the complete texts of the Herculaneum papyri, but will do so to a standard that exceeds current imagination.
We will wind back the clock on these manuscripts, generating images of their texts as if they were photographed the day they were written.

\section{Acknowledgements}

This work would be impossible without the tremendous contributions of many collaborators, over a long period and across many disciplines.
We thank multiple papyrologists, instrumental in providing feedback, advice, and interpretative contributions: 
James Brusuelas,
Tobias Reinhardt,
Gianluca Del Mastro,
Federica Nicolardi,
Richard Janko,
Robert Fowler,
Roger Macfarlane,
Daniel Delattre,
and Marzia D'Angelo.
Many students at the University of Kentucky have contributed to tool development and data processing, including
Bruno Athié-Teruel,
Jack Bandy,
Sydney Chapman,
Daniel Dopp,
and Kendall Weihe.
We are grateful to beamline scientists for their help with acquiring synchrotron CT data:
Jens Dopke at Rutherford Appleton Labs, and
Michael Drakopolous, Thomas Connolley, Robert Atwood, Leigh Connor, and Nghia Vo at the Diamond Light Source.
Our partners at the institutions hosting Herculaneum papyri have graciously offered material access, assistance during imaging sessions, and many other valuable contributions:
M. Zink,
M. Pastoureau,
M. Leclant,
M. Grimal,
M. Danesi,
M. Sabrina Castandet-Le Bris,
M. Fabien Queroux at the Institut de France;
and Francesco Mercurio,
Fabrizio Diozzi,
Luigi Vallefuoco,
and many others at the Biblioteca Nazionale di Napoli.
We are grateful for imaging resources and support from Micro Photonics, Graziano Ranocchia, Alessia Cedola,
Sara Stabile, Francesca Palermo, Inna Bukreeva, Daniela Mele,
and the UCLA School of Dentistry.
Infrared images are a result of much work done by Brigham Young University.
We thank the University of Kentucky's Center for Computational Sciences and Information Technology Services Research Computing for use of the Lipscomb Computing Cluster resources.
We gratefully acknowledge funding from The Andrew W. Mellon Foundation, The National Endowment for the Humanities, and the National Science Foundation.
W.B.S. acknowledges funding from the NSF (award IIS-1422039).
Any opinions, findings, and conclusions or recommendations expressed in this material are those of the author(s) and do not necessarily reflect the views of the NSF.
W.B.S. acknowledges funding from Google as well as individual supporters of EduceLab and the Digital Restoration Initiative at the University of Kentucky.

{\small
\bibliographystyle{ieee_fullname}
\bibliography{refs}

\begin{thebibliography}{10}
\bibitem{60minutes}
Can technology unravel the secrets sealed by mt. vesuvius 2,000 years ago?
\end{thebibliography}
}

\newpage
\appendix
\section{Methods}\label{methods}

The data in EduceLab-Scrolls is derived from two forms of imaging.
Volumetric X-ray CT is used to image all objects: both intact scrolls, and detached scroll fragments.
For scroll fragments with exposed text, spectral imaging is also used to capture the presence of ink on the surface.
To create a labeled dataset, a geometric software pipeline segments the surface in CT and then aligns the two image modalities.
Ink detection models can then be trained to learn the ink presence in the X-ray CT images.
Finally, the trained models are applied to hidden, subsurface layers in order to reveal previously unseen ink.
The following section describes these steps in higher detail.
For further clarification, we refer the reader directly to the Volume Cartographer and ink-ID implementations, available on GitHub.

\subsection{Tomographic imaging and reconstruction} \label{sec:imaging}

All volumetric datasets in EduceLab-Scrolls were acquired using X-ray micro-computed tomography (micro-CT) during multiple imaging sessions between 2009 and 2019.
Due to advances in imaging technology, advances in our understanding of the materials, and the availability of both samples and equipment over this time period, materials were not imaged under the same conditions.
The data was acquired using a range of imaging configurations and vary most importantly in their X-ray source types, incident energies, and spatial resolutions.

Imaging and volumetric reconstruction for all datasets followed the same high level workflow.
X-ray projection images of a sample were captured from a dense number of rotational positions around the object, following a circular scan trajectory.
These projections were then reconstructed into tomographic volumes using some form of filtered backprojection \cite{hsieh2015fbp}, often employing an implementation provided alongside the scanning hardware.
Volumes were then sliced along the rotational axis and quantized with manually determined intensity windows to create 16-bit TIFF image stacks.

In circumstances where the object did not entirely fit into a single field-of-view, multiple overlapping offset images (horizontal and/or vertical) were acquired for each rotational position.
Horizontal offset images were stitched prior to reconstruction to form a single, wider projection image.
Images taken at vertical offset positions were reconstructed as separate CT volumes, which we refer to as ``slabs,'' and were later merged during post-processing to form a single, coherent volume.

\subsubsection{Institut de France, 2009} \label{sec:idf}

P.Herc.Paris.\ 3 and 4 were scanned at the Institut de France (Paris, France) in July 2009, marking the first time tomographic imaging was used to analyze the interior of closed Herculaneum scrolls.
Both scrolls were acquired using a Skyscan 1173 micro-CT machine which had been installed at the Institut solely for the duration of the imaging session.
This scanner was equipped with a Hamamatsu 40-130kV (8W) X-ray source with a $<5$\textmu{}m spot size and a 2240x2240 flat panel detector with a 50\textmu{}m pixel pitch.

Because the Skyscan 1173 scans objects in an upright orientation, custom-fitted scan cases were required for both scrolls in order to safely support each scroll during micro-CT acquisition.
Prior to the scanning session, the surface profile of both scrolls was acquired using a portable laser scanning system.
Since Herculaneum materials have low reflectivity in the visible light spectrum, many gaps in the laser-derived profile data had to be manually interpolated in order to construct a complete surface profile.
This final profile information was then used as the base model to which a case was molded using expanding urethane foam.

The projection images for both scrolls were captured at 70kV, 100\textmu{}A with an exposure time of 700ms and averaging over 8 consecutive frames.
Two horizontal and four vertical offset positions were used in both instances to capture the full length of each scroll.
For P.Herc.Paris.\ 4, this configuration produces a final reconstructed voxel size of 23.94\textmu{}m.
For P.Herc.Paris.\ 3, which has a larger diameter, the final reconstructed voxel size is 27.17\textmu{}m.
Both scans were reconstructed using the Skyscan NRecon software.

\subsubsection{Università degli Studi di Bari ``Aldo Moro'', 2018} \label{sec:bari}

In November 2018, an unnumbered (\textit{sine numero} or s.n.) fragment from cassetto 110 at the Officina dei Papiri Ercolanesi (Naples, Italy) was scanned multiple times at the Dipartimento Scienze della Terra e Geoambientali, Università degli Studi di Bari ``Aldo Moro'' (Bari, Italy).
The fragment was selected for imaging due to its compact size and clearly visible surface text.
For organizational purposes, we have labeled this fragment P.Herc.\ s.n.\ (Cass.\ 110).
In total, four scans of the fragment were captured using a Skyscan 1172 micro-CT machine equipped with a 20-100kV (10W) X-ray source with a $<5$\textmu{}m spot size and a 4000x2664 CCD detector with a 9\textmu{}m pixel pitch.

In order to maximize the possibility of differentiating the ink from the papyrus in the reconstructed data, the scans captured during this imaging session prioritized high resolutions ($<10$\textmu{}m voxel sizes) and low incident energies ($\le50$kV). 
The entire fragment was scanned across two incident energies at approximately 5 \textmu{}m voxel sizes.
The first of these was captured at 44kV, 222\textmu{}A with an exposure time of 2400ms.
The second scan was captured at 49kV, 200\textmu{}A with an exposure time of 2400ms and averaging over 5 consecutive frames.
One horizontal offset was used in both instances to achieve a final reconstructed voxel size of 5.09\textmu{}m and 4.96\textmu{}m respectively.

Two additional ``core scans'' were also performed to capture a central region of the fragment under alternative configurations.
Core scanning is often employed when it is impossible or infeasible to scan the entire sample at a specific resolution, instead targeting a small, representative region-of-interest.
As a preliminary to a much higher resolution scan, a 23mm diameter core scan was taken at 60kV, 165\textmu{}A with an exposure time of 2120ms and averaging over 5 consecutive frames.
Pixel binning (2x2) was employed to improve image exposure times, resulting in a final reconstructed voxel size of 11.95\textmu{}m.
An 8mm diameter core scan was also captured at 44kV, 222\textmu{}A with an exposure time of 2750ms and averaging over 3 consecutive frames.
Due to time constraints, offset imaging was not employed, resulting in a final reconstructed voxel size of 2.24\textmu{}m.

All scans captured during this session were reconstructed using the Skyscan NRecon software.

\subsubsection{UCLA School of Dentistry, 2019} \label{sec:ucla}

In June 2019, two intact scrolls, P.Herc.\ 803 and 804, were scanned at the UCLA School of Dentistry (Los Angeles, USA).
The scan session was organized in collaboration with the J.\ Paul Getty Museum, which subsequently presented both scrolls as part of the ``Buried by Vesuvius: Treasures from the Villa dei Papiri'' exhibition at the Getty Villa.
Both scrolls were acquired using a Skyscan 1173 micro-CT machine provided by the UCLA School of Dentistry.

Prior to the scanning session, the surface profile of both scrolls was used to construct custom-fitted scan cases in a process similar to the one used in 2009.
To avoid previous issues with gaps in a laser-derived surface profile, photogrammetry capture and reconstruction was instead used to generate the surface profile.
By augmenting image capture with external lighting, the surface features of both scrolls were very clear in the photogrammetry image set, and the surface profiles for each scroll could be reliably reconstructed with little manual intervention.
These surface profiles were loaded into CAD software and used to design custom 3D case models which were subsequently fabricated using selective laser sintering (SLS) of Nylon 12 powder.
To protect the scrolls from the possibility of unintended abrasive damage, each case was lined with a layer of Teflon Relic Wrap prior to sample mounting.

The projection images for both scrolls were captured at 70kV, 114\textmu{}A with an exposure time of 1115ms and averaging over 7 consecutive frames.
Two horizontal and three vertical offset positions were used to capture the majority of P.Herc.\ 803.
Two horizontal and four vertical offsets were used to capture the entirety of P.Herc.\ 804.
For both scrolls, this configuration produces a final reconstructed voxel size of 23.88\textmu{}m.
Both scans were reconstructed using the Skyscan NRecon software.

\subsubsection{Diamond Light Source, 2019} \label{sec:diamond}

In September 2019, two scrolls (P.Herc.Paris.\ 3 and 4) and four fragments (P.Herc.Paris.\ 1 fr.\ 34, P.Herc.Paris.\ 1 fr.\ 39, P.Herc.Paris.\ 2 fr.\ 47, and P.Herc.Paris.\ 2 fr.\ 143) were scanned over many days on the I12 beamline \cite{drakopoulos2015i12} at the Diamond Light Source (Didcot, Oxfordshire, England).
Unlike the scanning environments used for the other datasets in this collection, the I12 beamline is an extremely brilliant light source which provides a parallel beam and monochromatic x-rays.
Though the parallel geometry does not support geometric magnification of the scan volume, four high-resolution optical modules provide pixel sizes from 18.53\textmu{}m down to 1.3\textmu{}m.
These system properties enabled the acquisition of extremely high-resolution scans of full scrolls and fragments that would have otherwise been infeasible in commercial micro-CT hardware.

Prior to the scanning session, scan cases were constructed for each scroll using the same method employed in \ref{sec:ucla}.
However, a manufacturing defect in the case for P.Herc.Paris.\ 3 resulted in that sample being scanned in the case constructed for the 2009 scan session (\ref{sec:idf}).
The fragments had each been previously mounted to tissue paper for conservation and storage.
For scanning, these were individually mounted into acrylic frames which held the tissue paper firmly in place while leaving the fragment undamaged.
These frames were then mounted to the sample stage with standard hardware.

Though the I12 sample stage provides motor-controlled vertical sample translation, the lengths of both scrolls exceeded the total vertical travel distance of the system.
To capture the full lengths of the scrolls, a manual lab jack was used to provide additional travel.
With the jack extended, each scroll was first scanned from the bottom to the furthest vertical position possible given the available motorized travel.
The jack was then compressed and the scroll was rescanned, this time from a point partway up the scroll to the top of the scroll.
This process produced two independent scans for each scroll, a lower and upper half which we have labeled Half 1 and Half 2 respectively.

The projection images of P.Herc.Paris.\ 3 and 4 were captured using optical module 2 at 54keV with an exposure time of 3ms.
Frame averaging was applied during flatfield acquisition but was not employed during primary imaging.
For P.Herc.Paris.\ 3, three horizontal and 11 vertical offsets were captured for Half 1, and three horizontal and five vertical offsets were captured for Half 2.
For P.Herc.Paris.\ 4, two horizontal and 11 vertical offsets were captured for Half 1, and two horizontal and 10 vertical offsets were captured for Half 2.
Due to a miscalculation in the projected position of the axis of rotation, a cylinder of diameter $\approx4.8$mm at the core of P.Herc.Paris.\ 3 is not represented in the projection images.
This results in a noticeable gap and reconstruction artifacts in the reconstructed volumes.
The final reconstructed voxel size for P.Herc.Paris.\ 3 and 4 is $\approx7.91$\textmu{}m.

The projection images of the four fragments were captured using optical module 3 at both 54keV and 88keV with an exposure time of 3ms.
The number of offsets used for each fragment was selected in order to fully capture each sample, and the offset positions were fixed between the 55keV and 88keV scans.
Two horizontal and five vertical offsets were captured for P.Herc.Paris.\ 1 fr.\ 34.
Two horizontal and seven vertical offsets were captured for P.Herc.Paris.\ 1 fr.\ 39.
Two horizontal and six vertical offsets were captured for P.Herc.Paris.\ 2 fr.\ 47.
Three horizontal and 10 vertical offsets were captured for P.Herc.Paris.\ 2 fr.\ 143.
The final reconstructed voxel size for all fragment scans is $\approx3.24$\textmu{}m.

All scans captured during this session were reconstructed using the Savu tomography reconstruction and processing pipeline \cite{savu2015atwood}.

\subsection{Spectral surface imaging}

The infrared surface images were acquired by Brigham Young University (BYU) using a multispectral imaging (MSI) system beginning in 1999.
Images were acquired across a range of wavelengths, from 450nm to 1000nm.
Due to the varying dimensions of Herculaneum scroll fragments, the resolution ranges from 300 to 600 dpi.
More details can be found in the original publications about this imaging work \cite{booras2001herculaneum,ware2000multispectral}.

\begin{figure}
    \centering
    \begin{subfigure}[b]{0.45\linewidth}
        \includegraphics[width=\linewidth]{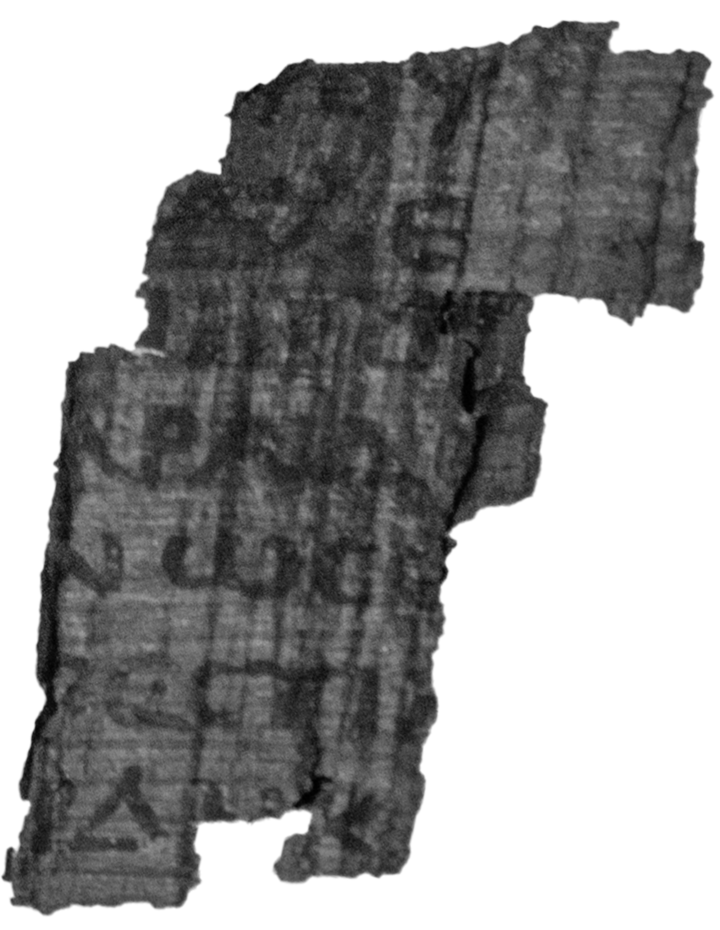}
        \caption{RGB.}
    \end{subfigure}
    \begin{subfigure}[b]{0.45\linewidth}
        \includegraphics[width=\linewidth]{figures/fragment_results/f1_ir.png}
        \caption{1000nm infrared.}
    \end{subfigure}
    \caption{
        Visible wavelength RGB and 1000nm infrared images of P.Herc.Paris.\ 2 fr.\ 47, revealing improved ink contrast in infrared.
    }
    \label{fig:rgb-alongside-ir}
\end{figure}

The 1000nm images from the BYU dataset, and those from similar wavelengths such as 950nm, have become standard resources for papyrologists working with Herculaneum papyri due to the high contrast captured between ink and papyrus.
This contrast dramatically exceeds that observed by the naked eye in visible wavelengths, where the black ink can be difficult to see against the carbonized papyrus background.
For this reason, 1000nm images are chosen as the ground truth images in our work.
Figure \ref{fig:rgb-alongside-ir} illustrates the improved contrast achieved with spectral imaging.

\subsection{Segmentation}\label{segmentation}

The creation of a labeled training dataset relies on registering, or aligning, the 3D CT and 2D infrared images into the same geometric space.
This registration creates aligned image to image labels, enabling supervised learning so that models can be trained to detect the ink presence in CT.
There are multiple steps in this process, but the purpose throughout is simple, and is to achieve the above alignment.
The following describes the process for one scroll fragment (P.Herc.Paris.\ 2 fr.\ 47), which was repeated for each fragment in EduceLab-Scrolls.
The first step is the segmentation of the fragment surface from the volumetric CT image, resulting in a 3D mesh closely following the fragment's exterior surface shape.

\subsubsection{Initial segmentation with Canny edge detector}

\begin{figure}
    \centering
    \includegraphics[width=\linewidth]{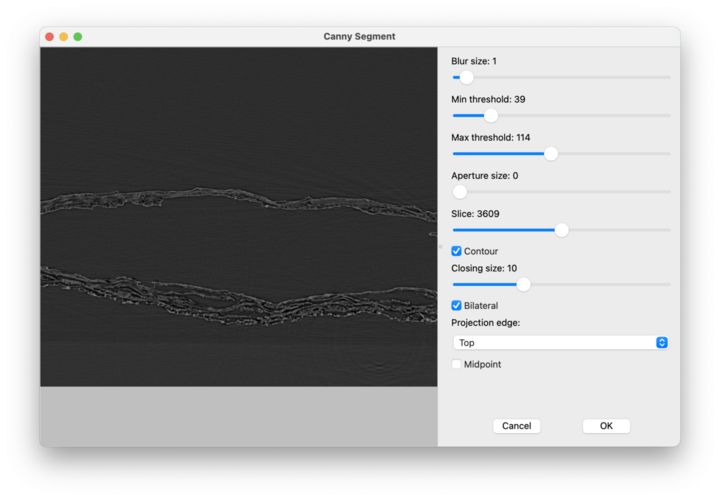}
    \caption{Volume Cartographer's Canny Segment application.}
    \label{fig:vc-canny-segment}
\end{figure}

An initial segmentation is constructed per-slice using a Canny edge detector \cite{canny1986computational}.
The output of this step is a dense 3D point cloud, with points defining the fragment surface in the coordinate space of the volumetric CT scan.
This approach is implemented in the Canny Segment application of Volume Cartographer \cite{parker2021vcsoftware}, our open source toolkit for virtual unwrapping.
Shown in Figure \ref{fig:vc-canny-segment}, an interactive graphical user interface (GUI) displays a slice of the volumetric image, along with the Canny edge detector output overlaid on the slice image.
A slider allows the user to move through the different volume slices, and a set of controls allows the user to tune the Canny edge detector parameters.
The user's objective is to tune the Canny parameters until the detected edges precisely follow the fragment surface in the slice image.
Empirically, different CT images can result in different optimal settings, so the tuning step is necessary for each dataset.
Eventually, this tuning could be automated, but the current human-in-the-loop approach is not a bottleneck in the overall pipeline.
Table \ref{table:canny-settings} shows the Canny parameters chosen for each fragment using this process.

\begin{table*}
    \centering
    \begin{tabular}{llllllll}
        \toprule
        & Blur & Min & Max & Aperture & Contour & Closing size & Bilateral \\
        \midrule
        P.Herc.Paris.\ 1 fr.\ 34, 54keV, surface & 1 & 39 & 114 & 0 & \checkmark & 10 & \checkmark \\
        P.Herc.Paris.\ 1 fr.\ 39, 54keV, surface & 1 & 39 & 114 & 0 & \checkmark & 10 & \checkmark \\
        P.Herc.Paris.\ 2 fr.\ 47, 54keV, surface & 1 & 39 & 114 & 0 & \checkmark & 10 & \checkmark \\
        P.Herc.Paris.\ 2 fr.\ 143, 54keV, surface & 1 & 39 & 114 & 0 & \checkmark & 16 & \checkmark \\
        P.Herc.Paris.\ 1 fr.\ 39, 54keV, hidden & 1 & 39 & 155 & 0 & & -- & \checkmark \\
        P.Herc.Paris.\ 2 fr.\ 47, 54keV, hidden & 1 & 76 & 45 & 0 & & -- & \checkmark \\
        P.Herc.Paris.\ 2 fr.\ 143, 54keV, hidden & 1 & 39 & 114 & 0 & & -- & \checkmark \\
        \bottomrule
    \end{tabular}
    \caption{Canny edge detection parameters chosen for each segmentation in EduceLab-Scrolls.}
    \label{table:canny-settings}
\end{table*}

Once the Canny edge detector parameters are tuned, the user closes the graphical application and the edge detection is applied to each volume slice in sequence using a procedure shown in Algorithm \ref{alg:canny} and discussed here.
For each slice, Canny edge detection is computed using the tuned parameters.
A set of rays is then projected into the slice image from an image boundary.
For each ray, if it intersects a detected edge, this intersection point is added to a 3D point cloud using the $(x, y)$ from the image pixel and $z$ from the slice index.
If the ray departs the image bounds before intersecting a detected edge, no point is added.

The image boundary (top, bottom, left, or right) used as the set of ray origins is configurable.
Since the CT volumes in EduceLab-Scrolls are oriented consistently, the top edge of the slice image was always used as the projection edge in this method.
This is reflected in Algorithm \ref{alg:canny}, which is fixed to use the top edge.
In this configuration, rays are projected downwards along each image column until intersecting the first detected edge.

\begin{algorithm}
    \caption{
        Surface segmentation using Canny edge detection.
        Given a volume $V$ and Canny edge detection parameters $C$, returns a dense point cloud $P$ with points defining the fragment surface.
        Implemented in \texttt{CannySegment} application of Volume Cartographer.
    }\label{alg:canny}
    \begin{algorithmic}[1]
        \Procedure{Canny-Segment}{$V,C$}
            \State $P = \emptyset$ \Comment empty point cloud
            \For{$z_i \in [0, V.\text{slices} - 1]$}
                \State $I_i$ = $V[z_i]$ \Comment extract slice image
                \State $E_i = \Call{Canny}{I_i, C}$ \Comment detect edges
                \For{$x_j \in [0, E_i.\text{cols} - 1]$}
                    \For{$y_k \in [0, E_i.\text{rows} - 1]$}
                        \If{\Call{Edge}{$E_i[y_k, x_j]$}}
                            \State $P = P \cup (x_j, y_k, z_i)$ \Comment add point
                            \State \textbf{break}
                        \EndIf
                    \EndFor
                \EndFor
            \EndFor
            \State \textbf{return} $P$
        \EndProcedure
    \end{algorithmic}
\end{algorithm}

\begin{figure}
    \centering
    \begin{subfigure}[b]{0.45\linewidth}
        \includegraphics[width=\linewidth]{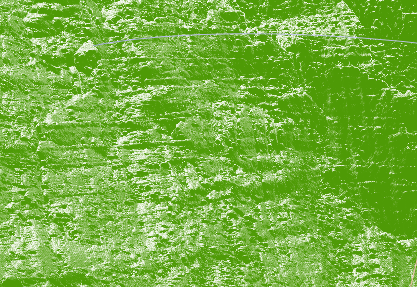}
        \caption{Papyrus, fragment surface.}
    \end{subfigure}
    \begin{subfigure}[b]{0.45\linewidth}
        \includegraphics[width=\linewidth]{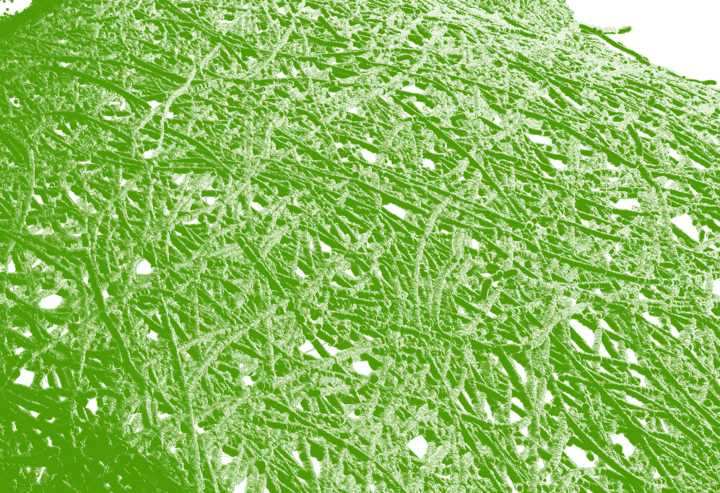}
        \caption{Paper, backing sheet.}
    \end{subfigure}
    \caption{
        Details of raw point cloud, showing surface textures recovered using Canny segmentation.
        Visualized using Meshlab.
    }
    \label{fig:point-clouds}
\end{figure}

The raw point clouds generated using this method contain a high level of detail.
Visualization reveals not only the general shape of the surface, but also textural details.
Figure \ref{fig:point-clouds} shows two examples of this, where the grid-like structure of the papyrus fibers is visible on the fragment, and the paper fibers are visible on the backing sheet to which the fragment is mounted.

\subsubsection{Mesh processing}\label{mesh-processing}

The point cloud generated using the Canny segmentation method is dense, and can contain on the order of 10-100M points.
This number of points is difficult to visualize interactively or process effectively, and contains more detail than is necessary for a precise surface segmentation.
Therefore, the first step in mesh processing is to simplify the raw point cloud.
We have found empirically that simplifying the point cloud to $\approx$100,000 points preserves enough detail while easing downstream steps.
Point cloud simplification is performed using the open source Meshlab interface \cite{LocalChapterEvents:ItalChap:ItalianChapConf2008:129-136}, as are the remainder of the steps in \ref{mesh-processing}.

\begin{figure}
    \centering
    \begin{subfigure}[b]{0.45\linewidth}
        \includegraphics[width=\linewidth]{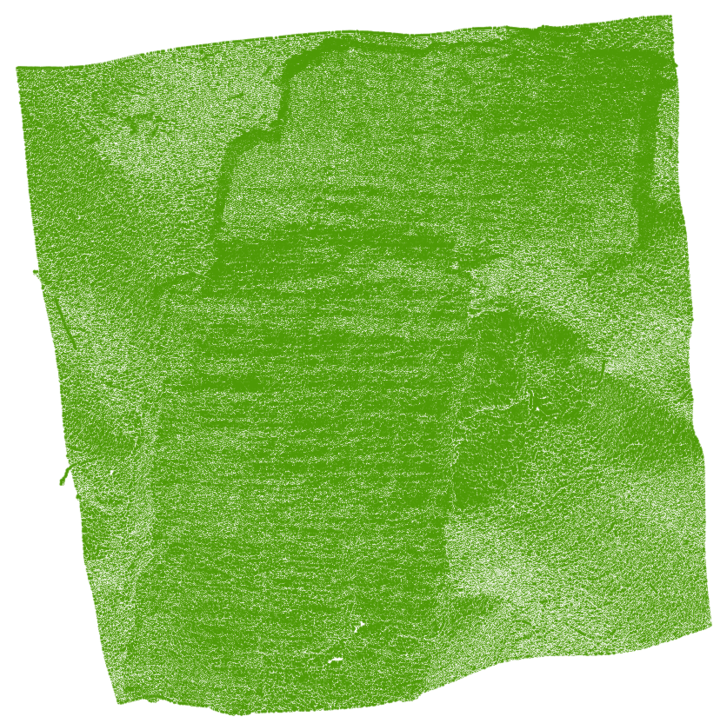}
        \caption{Simplified point cloud.}
    \end{subfigure}
    \begin{subfigure}[b]{0.45\linewidth}
        \includegraphics[width=\linewidth]{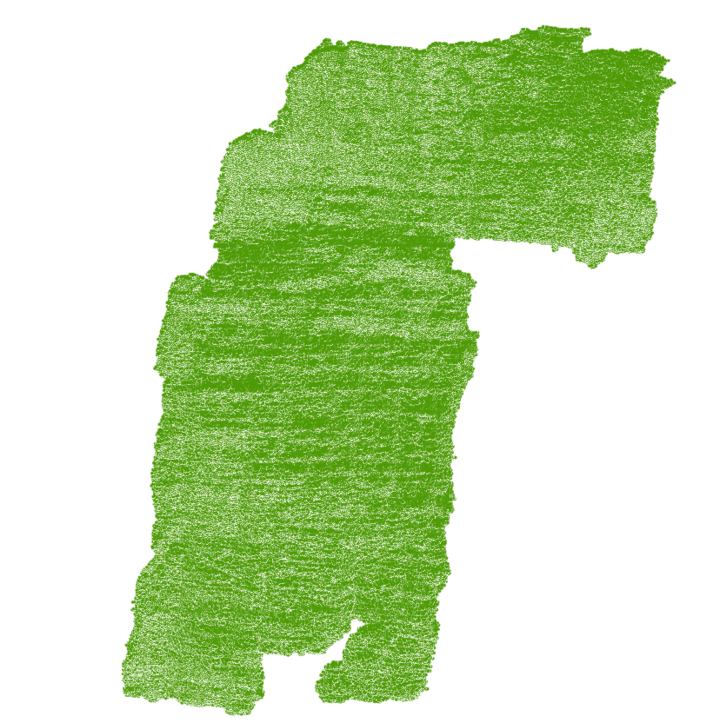}
        \caption{Cleaned point cloud.}
    \end{subfigure}
    \caption{
        P.Herc.Paris.\ 2 fr.\ 47 point cloud before and after the manual removal of extraneous points.
        Visualized using Meshlab.
    }
    \label{fig:point-cloud-cleaning}
\end{figure}

The point cloud also captures features from the CT volume that do not lie along the fragment surface.
Typically these are other physical structures present in the scan volume, such as the backing paper upon which the papyrus fragment is mounted.
Sometimes other small features appear due to artifacts in the CT images.
These are removed manually, by selecting and deleting the extraneous points.
Figure \ref{fig:point-cloud-cleaning} shows the point cloud before and after this step.

At this stage, the surface typically looks clean when viewed from above, but can be noisier below the surface.
This is an artifact of the Canny segmentation method described above.
Even with fine-tuned parameters, the edge detection often has some false negatives, where in small regions an edge is not detected even though the surface boundary passes there.
In these locations, the projected rays in the segmentation algorithm proceed below the true surface, intersecting another detected edge somewhere beneath and adding that point to the cloud.
This slight preference for false negatives is deliberately chosen during the Canny parameter tuning, as false positives create a noisier surface from \textit{both} sides when they detect edges in the CT image noise in regions of air.

This noise can bias the subsequent meshing step, pulling the segmentation down into the papyrus instead of the desired behavior which precisely follows the boundary between papyrus and air.
To alleviate this artifact, outlier points are selected and deleted.
Interactive visualization during the outlier selection allows one to tune the outlier detection until the intended points are selected.
This selection and deletion is often performed two or three times.
Sometimes, excellent performance of the initial Canny segmentation can render outlier removal unnecessary.
This is typically the result of enabling the \texttt{--contour} option in Canny segmentation, for those datasets which are cooperative.

\begin{figure}
    \centering
    \begin{subfigure}[b]{0.45\linewidth}
        \includegraphics[width=\linewidth]{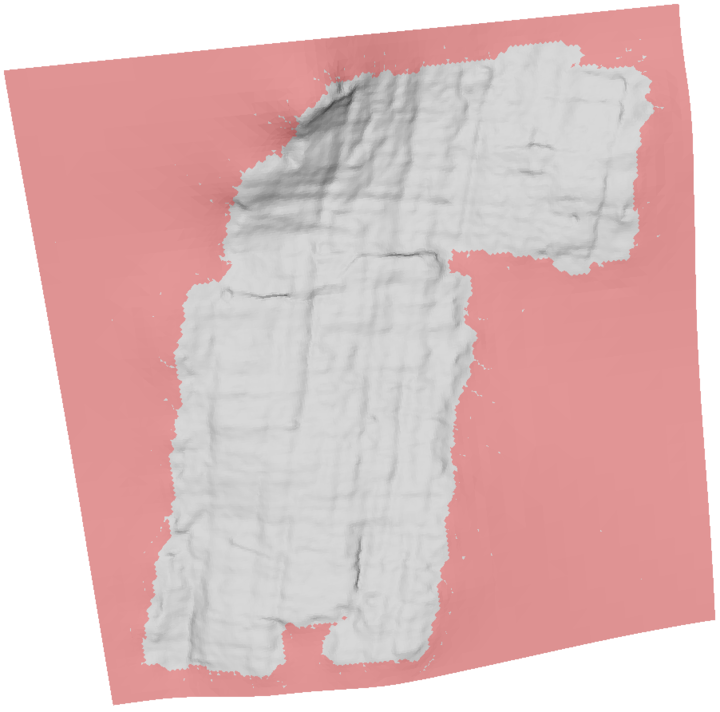}
        \caption{Initial reconstructed surface, with faces selected for deletion.}
    \end{subfigure}
    \begin{subfigure}[b]{0.45\linewidth}
        \includegraphics[width=\linewidth]{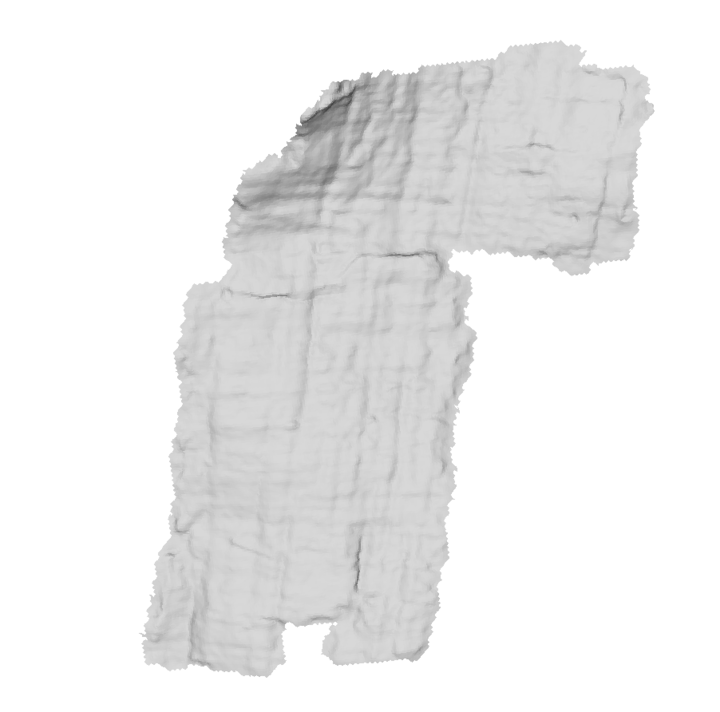}
        \caption{Final mesh after cleaning steps.}
    \end{subfigure}
    \caption{
        P.Herc.Paris.\ 2 fr.\ 47 segmented mesh before and after final cleaning.
        Visualized using Meshlab.
    }
    \label{fig:mesh-cleaning}
\end{figure}

The processed point cloud is now meshed using screened Poisson surface reconstruction \cite{kazhdan2013screened}.
As this reconstruction produces a full sheet that exceeds the fragment bounds, extraneous faces are selected and removed (Figure \ref{fig:mesh-cleaning}).
This selection is either performed by selecting faces with a thresholded Hausdorff distance \cite{cignoni1998metro} between the mesh and source point cloud, or by selecting faces with an edge length above some threshold.
A final cleaning is then performed, consolidating the mesh to a single connected component, closing small holes, and ensuring the mesh is two-manifold.

\begin{figure}
    \centering
    \includegraphics[width=\linewidth]{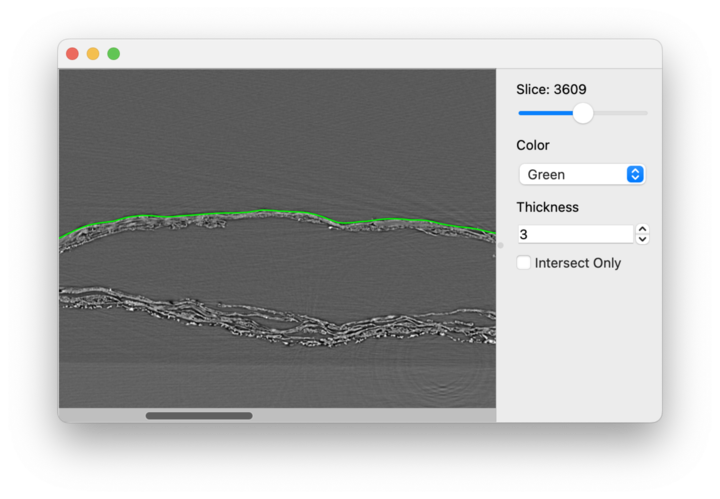}
    \caption{
        Intersection of the final segmented surface mesh with a slice from the original CT volume.
        Visualized using the Projection application of Volume Cartographer.
    }
    \label{fig:projection}
\end{figure}

The resulting mesh precisely follows the exposed surface of the fragment.
The 3D surface now resembles the general shape of the fragment as seen in a photograph, and the surface still captures details such as the papyrus fiber structure.
The segmentation can be validated by viewing the intersection of the final mesh with the slice images of the original volume (Figure \ref{fig:projection}).

\subsection{Flattening}

Recall that the purpose of this multi-step process is to create an image space to which both the volumetric CT data and the 2D surface images can be aligned.
This space is created in this step, by flattening the segmented surface mesh to a 2D image plane.
The resulting mesh parameterization defines a bijective mapping, in which points on the mesh surface can be translated between volumetric CT coordinates and 2D image coordinates.

There are various methods to define this mapping between 3D mesh and image plane.
Traditionally, the objective of virtual unwrapping is to ``unwrap'' or ``flatten'' a convoluted surface so that the text can be more easily read.
In that context, the parameterization prioritizes angle preservation in order to avoid distortion.
Approaches such as angle based flattening (ABF) \cite{sheffer2001parameterization} and least squares conformal mapping (LSCM) \cite{levy2002least} are based on this principle.

For the scroll fragments to be used as training data, the objective is slightly different.
The mapping will generate a 2D image to which the spectral photographs will later be aligned.
To ease the alignment and minimize the deformations necessary, an 
orthographic projection is used instead of an angle preserving method.

To compute the orthographic projection, an oriented bounding box is first computed around the 3D mesh.
The longest and second longest axes of the bounding box are chosen to define the 2D image plane.
The mesh points are projected along the third bounding box axis onto this plane, defining the mapping.
Algorithm \ref{alg:orthographic} summarizes this projection method, which is implemented in Volume Cartographer along with ABF and LSCM.

\begin{algorithm}
    \caption{
        The orthographic projection method for mesh flattening.
        Given a mesh $M$, computes a UV map $U$ which contains $(u_p, v_p)$ for each vertex $\textbf{p} \in M$, mapping the 3D point to the 2D image plane.
        Implemented in \texttt{OrthographicProjectionFlattening} class of Volume Cartographer.
    }\label{alg:orthographic}
    \begin{algorithmic}[1]
        \Procedure{Orthographic-Projection}{$M$}
            \State $B = \Call{Oriented-Bounding-Box}{M}$
            \State $l_u = \Call{Norm}{B.\text{axes}[0]}$ \Comment longest axis
            \State $l_v = \Call{Norm}{B.\text{axes}[1]}$ \Comment second longest axis
            \State $\textbf{u} = B.\text{axes}[0] / l_u$
            \State $\textbf{v} = B.\text{axes}[1] / l_v$
            \State $\textbf{o} = B.\text{origin}$
            \State $U = []$ \Comment empty UV map
            \For{$\textbf{p} \in M$} \Comment for each mesh point
                \State $u_p = (\textbf{p} - \textbf{o}) \cdot \textbf{u}$ \Comment project to image plane
                \State $v_p = (\textbf{p} - \textbf{o}) \cdot \textbf{v}$
                \State $u_p = u_p / l_u$ \Comment normalize UV to $[0,1]$
                \State $v_p = v_p / l_v$
                \State $U.\text{append}((u_p, v_p))$ \Comment add to UV map
            \EndFor
            \State \textbf{return} $U$
        \EndProcedure
    \end{algorithmic}
\end{algorithm}

\subsection{Per-pixel map}

The mapping between image spaces is bidirectional; that is, we should be able to map mesh points in the 3D volume space to the 2D flattened image space and vice versa.
The UV map computed with the orthographic projection explicitly defines the 3D $\rightarrow$ 2D mapping from the mesh to the flattened image plane.
The 2D $\rightarrow$ 3D mapping is implicitly defined, but it is helpful in later steps to store this explicitly as well.
A per-pixel map (PPM) is the file used to represent this mapping.
The PPM is effectively a 6-channel image file sharing the $(x, y)$ dimensions of the flattened image space.
For each pixel $(x_i, y_j)$, the PPM stores the 3D mesh position $(v^{ij}_x, v^{ij}_y, v^{ij}_z)$ and the mesh surface normal vector $(n^{ij}_x, n^{ij}_y, n^{ij}_z)$ at that location.
For brevity these vectors will be notated $(v_x, v_y, v_z)$ and $(n_x, n_y, n_z)$, and sometimes $\textbf{p}$ and $\textbf{n}$ respectively.

The PPM is computed by iterating over the 2D pixels, mapping each one to its mesh triangle (if one exists) based on the UV mapping, and using the barycentric triangle coordinates to interpolate the surface position and normal vector (normalized) between the triangle vertices.
This method is implemented in Volume Cartographer, and is not specific to orthographic projection, functioning identically with the other flattening methods.

\begin{figure}
    \centering
    \begin{subfigure}[b]{0.3\linewidth}
        \includegraphics[width=\linewidth]{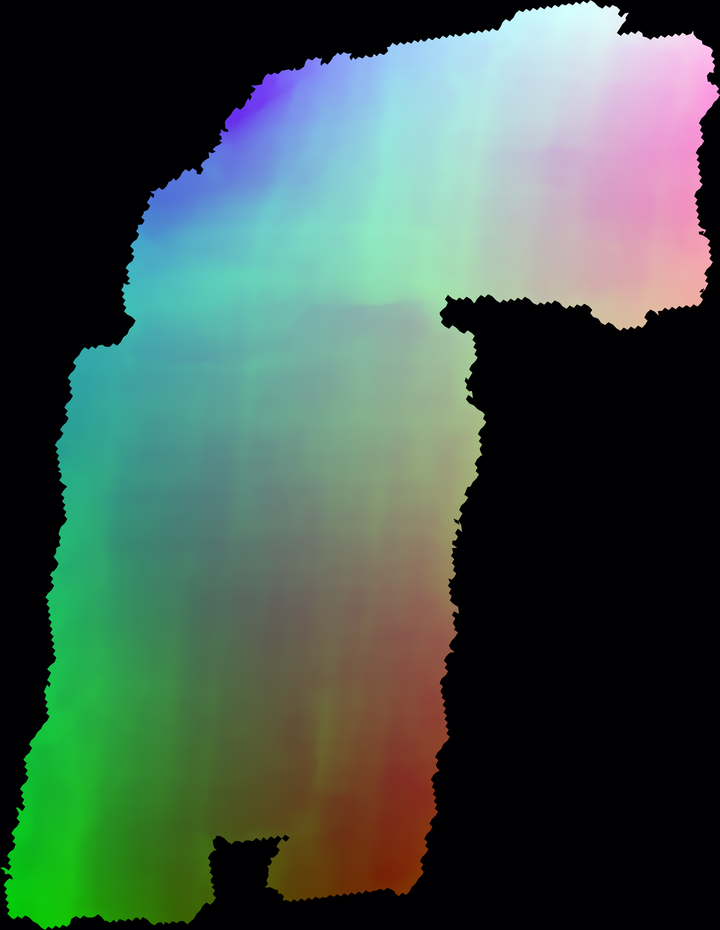}
        \caption{3D position.}
    \end{subfigure}
    \begin{subfigure}[b]{0.3\linewidth}
        \includegraphics[width=\linewidth]{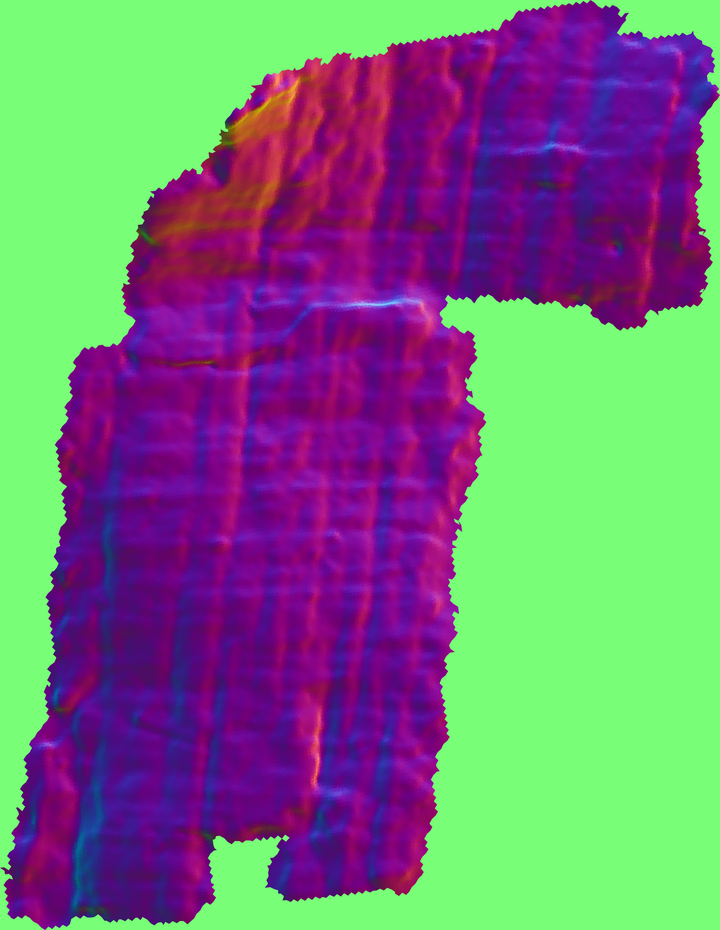}
        \caption{Normal vector.}
    \end{subfigure}
    \begin{subfigure}[b]{0.3\linewidth}
        \includegraphics[width=\linewidth]{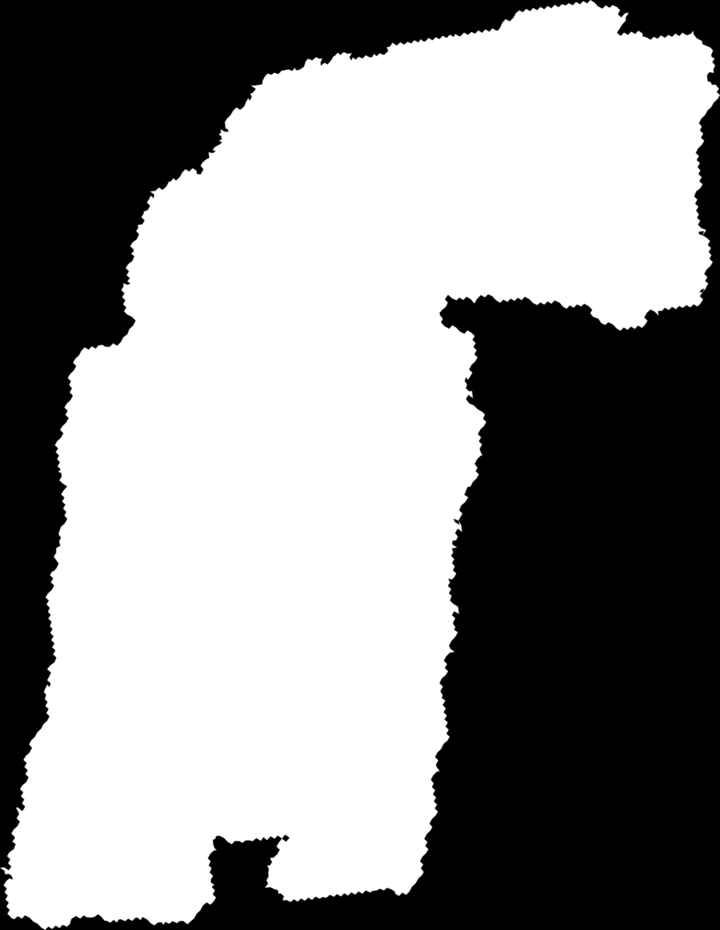}
        \caption{Surface mask.}
    \end{subfigure}
    \caption{
        Visualization of P.Herc.Paris.\ 2 fr.\ 47 PPM data.
        (a) 3D position $(v_x, v_y, v_z)$ mapped to RGB channels.
        (b) Normalized surface normal vector $(n_x, n_y, n_z)$ mapped to RGB.
        (c) Binary mask indicating surface bounds.
    }
    \label{fig:ppm}
\end{figure}

Figure \ref{fig:ppm} visualizes the PPM data for P.Herc.Paris.\ 2 fr.\ 47 by mapping the 3D coordinates and the surface normal vector values to RGB channels in a color image.
Notably, the visualization of surface normals reveals some texture on the surface, showing the grid-like structure of papyrus fibers.

\subsection{Surface volume}

Prior work for machine-learning based ink detection \cite{parker2019inkid} has used the PPM as the primary image space during training and inference.
Inputs are sampled from CT by first sampling $(x_i, y_j)$ from the PPM.
The 3D position $(v_x, v_y, v_z)$ and surface normal vector $(n_x, n_y, n_z)$ are read from the PPM, and used to sample a local neighborhood called a subvolume from the CT volume.

This approach is flexible but has a few drawbacks.
First, as implemented in ink-ID, the entire CT volume for each dataset had to be stored in memory during training or inference.
It is often desirable to train or predict across multiple labeled scroll fragments simultaneously, but the CT volumes are large enough that they limited experimentation to a single fragment at a time using the available hardware.

Second, while the PPM defines a mapping between the 2D and 3D image spaces, the CT inputs and surface labels are not themselves registered (aligned) to one another.
Processing them jointly relies on the pointwise mapping defined by the PPM, and there is not a natural way to use other sampling methods common to image-to-image domains such as aligned image patches.

Finally, the PPM itself is often multiple GB, a nontrivial increase to the already tight memory requirements.

To alleviate these pain points, this work samples ``surface volumes'' from the original CT volumes, based on the segmented surface meshes.
The surface volume contains the CT data intersected by a segmented mesh, as well as the neighboring CT data within a configurable thickness.
The surface volume is sampled such that its X-Y slices are inherently aligned with the PPM image.
The aligned image labels (see \ref{alignment}) will therefore also be registered to the CT images directly.
The PPM can then be discarded, having served its purpose.
Figure \ref{fig:method-overview} illustrates the objective of this registration, where \ref{fig:overview-d}, \ref{fig:overview-f}, and \ref{fig:overview-g} are aligned to each other in X-Y.

Algorithm \ref{alg:surface-volume} shows the surface volume sampling method.
The algorithm iterates over the PPM $P$, and for each point reads the PPM at that $(x, y)$.
This provides the 3D coordinate $\textbf{p}$ and normal vector $\textbf{n}$ of the same point in 3D, which are used to sample a linear neighborhood about that point in 3D using trilinear interpolation.
This line fills the depth of the surface volume $S$ for one $(x, y)$.
When repeated for each pixel, the result is a surface volume containing the CT data surrounding the segmented mesh.

\begin{algorithm}
    \caption{
        Surface volume generation.
        Given a PPM $P$, CT volume $V$, sampling radius $r \in \mathbb{N}$, and interval $\delta \in \mathbb{R}_{>0}$, returns the sampled surface volume $S$.
        Implemented in \texttt{vc\_layers\_from\_ppm} utility of Volume Cartographer.
    }\label{alg:surface-volume}
    \begin{algorithmic}[1]
        \Procedure{Surface-Volume}{$P, V, r, \delta$}
            \State $d = 2r + 1$ \Comment surface volume depth
            \State $S = \Call{Zeros}{(d, P.\text{rows}, P.\text{cols})}$
            \For{$y_i \in [0, P.\text{rows} - 1]$} \Comment iterate PPM X-Y
                \For{$x_j \in [0, P.\text{cols} - 1]$}
                    \State $(v_x, v_y, v_z, n_x, n_y, n_z) = P[y_i, x_j]$
                    \State $\textbf{p} = [v_x, v_y, v_z].\text{T}$ \Comment 3D coordinates
                    \State $\textbf{n} = [n_x, n_y, n_z].\text{T}$ \Comment normal vector
                    \For{$z_k \in [-r, r]$} \Comment iterate Z (depth)
                        \State $\hat{\textbf{p}} = \textbf{p} + z_k \delta \textbf{n}$ \Comment compute 3D position
                        \State $v = \Call{Interpolate}{V, \hat{\textbf{p}}}$ \Comment sample
                        \State $S[z_k + r, y_i, x_j] = v$ \Comment store
                    \EndFor
                \EndFor
            \EndFor
            \State \textbf{return} $S$
        \EndProcedure
    \end{algorithmic}
\end{algorithm}

For the surface volumes in this work, the sampling radius was set to $r=32$ (creating a surface volume of 65 channels) and the interval $\delta=1$ was chosen to preserve the original image resolution.
The surface volume for P.Herc.Paris.\ 2 fr.\ 47 is visualized in Figure \ref{fig:overview-d} and represents the input to the ink detection model during training and inference.

Generating surface volumes is a known compromise, as it introduces slight aliasing effects where there is curvature on the mesh surface.
In our experiments, this does not impact the model's ability to learn to identify the presence of ink.
There are other approaches that would alleviate the above pain points without introducing aliasing, such as storing volumes as 3D chunks on disk and using caching to limit the memory usage.
Our initial implementations of this method have been prohibitively slow, but this is a matter of engineering and we believe will be addressed in the near future.

\subsection{Texture image}

The steps so far have been concerned with creating the model inputs; that is, the CT surface volumes based on segmented surfaces.
Recall from \ref{segmentation} that the overarching objective of this pipeline is to create a registered image dataset, in which the CT volumes are paired with aligned labels.

We now transition to the label creation and alignment.
First, we will use the existing PPM to generate an image to which the label images can be registered.
By design, this image occupies precisely the same space as the UV map generated for the mesh.
For this reason, it can be used directly as a texture for mesh rendering, and so it is called a ``texture image''.
The texture image plots the CT intensity on the PPM image space, providing a visualization of the segmented surface as seen in X-ray.

Texture image generation is similar to generating a surface volume, in that we iterate across the PPM surface, read the local 3D position and normal vector, and then sample a local neighborhood from the original CT volume.
Instead of storing an entire linear neighborhood for each PPM pixel, the local neighborhood is filtered to produce a single intensity value which is then plotted on the texture image.
This filter could be any function, and there are multiple filters implemented in Volume Cartographer.

The texture images used in this work all sample a bidirectional linear neighborhood, centered on the point coordinates and oriented along the normal vector as read from the PPM.
The length of this neighborhood is set to equal the estimated material thickness, which is set to 150\textmu{}m for all texture images in this work.
The actual thickness of papyrus sheets varies considerably, but this has been an effective estimate for this purpose.

A max filter is then used, which simply takes the maximum value from this neighborhood and plots it.
The texturing methods used are consistent with prior work and more information can be found there \cite{seales2016engedi}.
Though the implementation differs, note that generating a texture image using a max filter is analogous to generating a surface volume of the same depth and then computing a max projection on the Z axis.

For those artifacts with clear X-ray contrast between ink and substrate, the texture image reveals the text and the pipeline is complete.
Of course, the lack of said contrast is the motivation behind the pipeline presented here.
Multiple simple filters have been tried in an attempt to extract visible ink contrast, so far without success. 
In this case, rather than revealing text, the function of the texture image is to serve as the fixed image for label image registration.

\begin{figure}
    \centering
    \begin{subfigure}[b]{0.45\linewidth}
        \includegraphics[width=\linewidth]{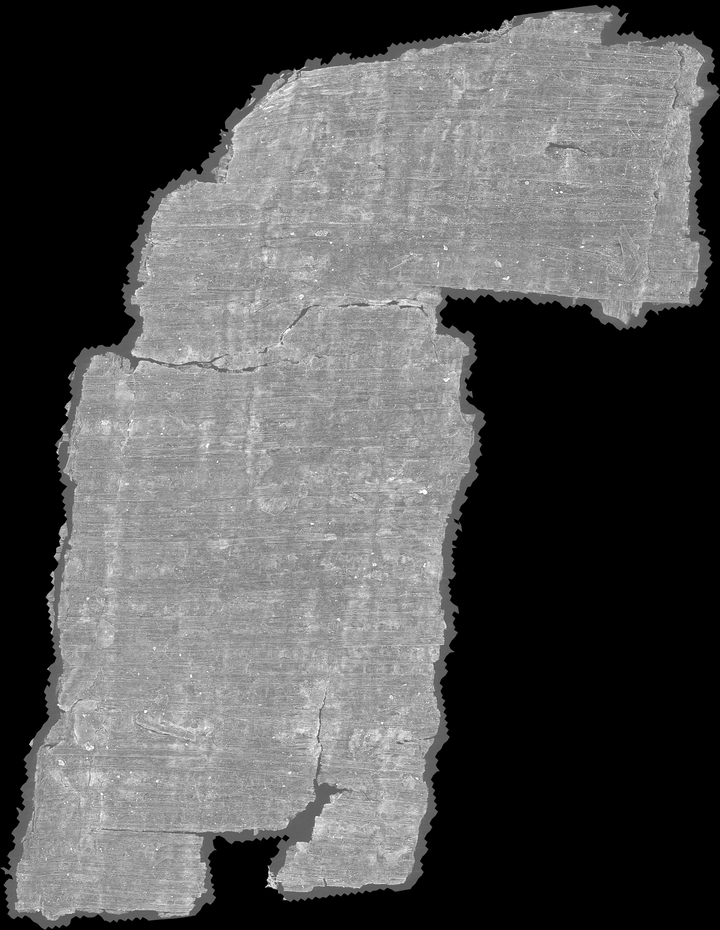}
        \caption{Texture image.}
    \end{subfigure}
    \begin{subfigure}[b]{0.45\linewidth}
        \includegraphics[width=\linewidth]{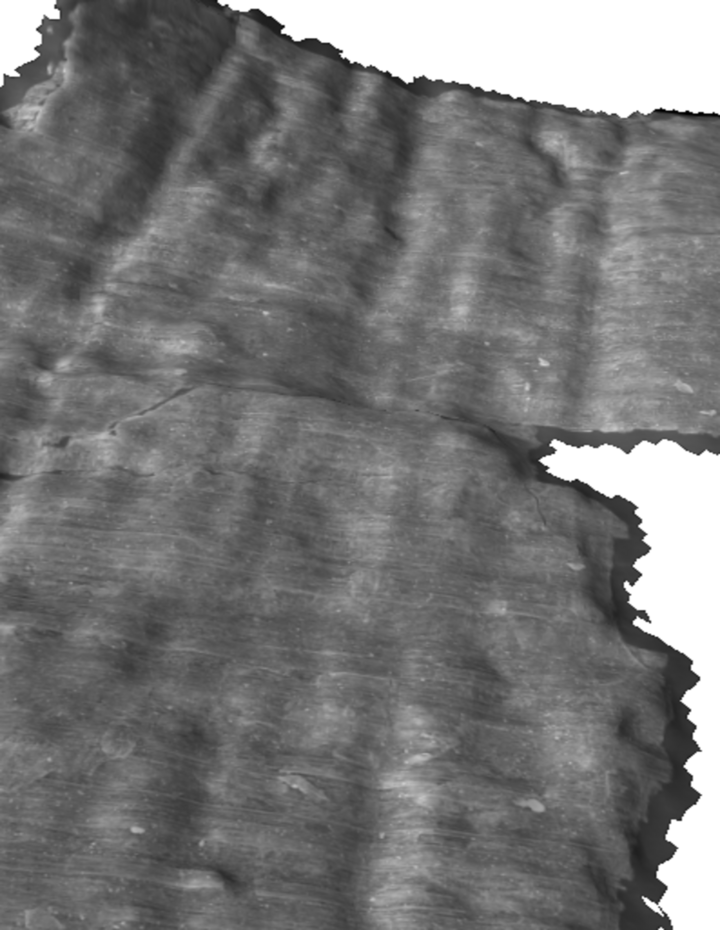}
        \caption{Textured mesh (detail).}
    \end{subfigure}
    \caption{
        Texture image.
    }
    \label{fig:texture}
\end{figure}

Figure \ref{fig:texture} shows the texture image generated for P.Herc.Paris.\ 2 fr.\ 47.
As mentioned, this can be used directly as a mesh texture for 3D rendering, also shown.
Any of the images subsequently registered to the texture image can therefore also be utilized in 3D rendering using the existing UV map.

\subsection{Label alignment}\label{alignment}

With the texture image generated, it is now possible to align the label images, pairing labels with the CT data.
As mentioned, infrared photographs are selected as the label photographs for their increased ink contrast.
This problem of alignment, known as image registration, involves identifying one image as the ``fixed'' image and another as the ``moving'' image.
In this case, the texture image is fixed, as it is tied to the generated UV map, and the infrared image is moved.
A deformable warp (as opposed to a simpler transformation such as affine or projective) is necessary to accommodate the subtle differences in image geometry, as one was acquired using a lens and the other was generated by an orthographic projection from CT.

\begin{figure}
    \centering
    \includegraphics[width=\linewidth]{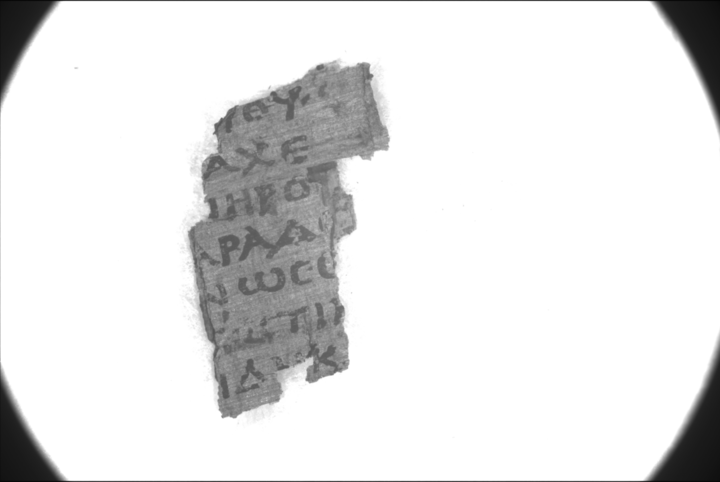}
    \caption{Infrared image before registration (alignment).}
    \label{fig:ir-original}
\end{figure}

Image registration is an established area of research, with many successful algorithms using feature- or intensity-based matching, in addition to other approaches \cite{sotiras2013deformable}.
Feature- and intensity-based methods both struggle with the multimodal nature of this image registration.
The ink appearing in one modality and not the other is a significant challenge.
In addition, those features that do appear in both images (typically papyrus fibers) have a significantly different appearance in each.
Due to these challenges, we have not yet discovered an automated method that is capable of successfully aligning the texture and infrared images.

\begin{figure}
    \centering
    \begin{subfigure}[b]{0.45\linewidth}
        \includegraphics[width=\linewidth]{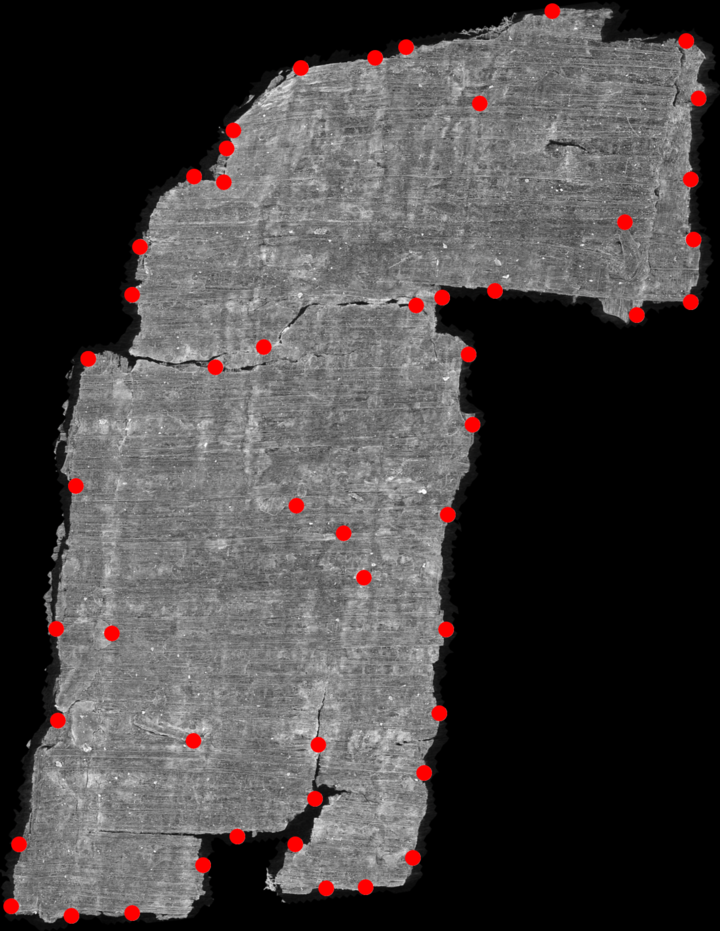}
        \caption{Texture image.}
    \end{subfigure}
    \begin{subfigure}[b]{0.45\linewidth}
        \includegraphics[width=\linewidth]{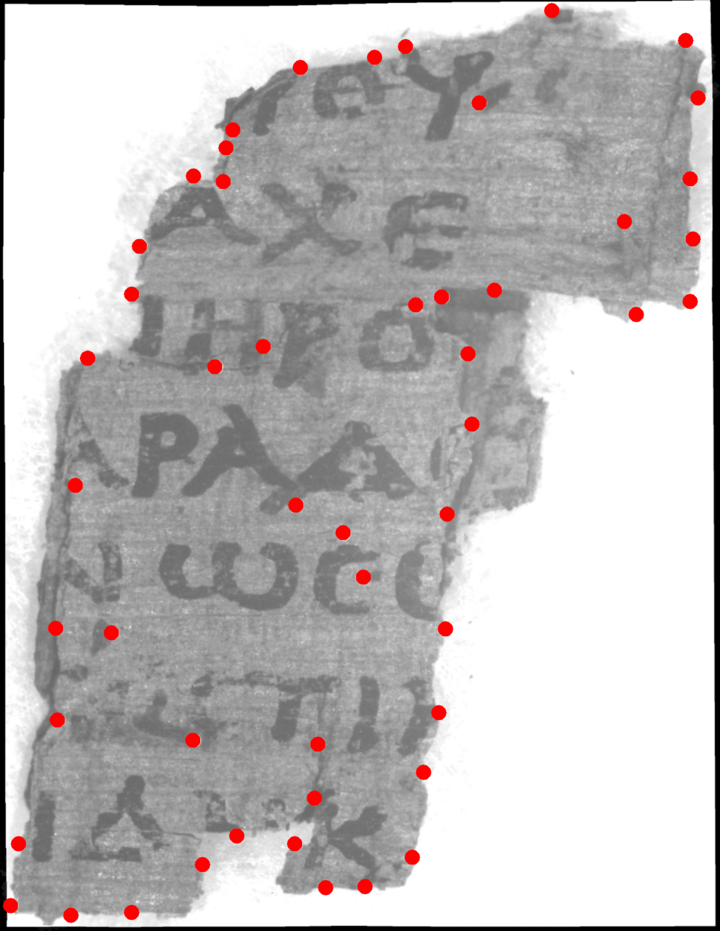}
        \caption{Infrared image.}
    \end{subfigure}
    \caption{
        Manually identified alignment points used in the registration process for P.Herc.Paris.\ 2 fr.\ 47.
    }
    \label{fig:alignment}
\end{figure}

\begin{figure}
    \centering
    \begin{tabular}{cc}
        \includegraphics[width=0.45\linewidth]{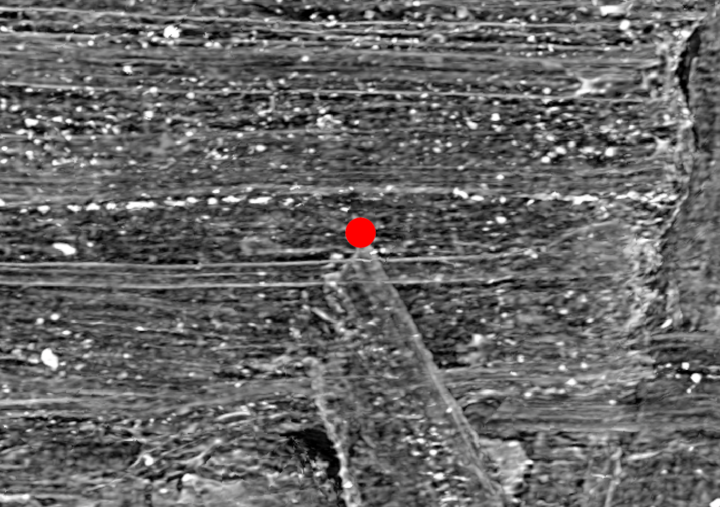} &
        \includegraphics[width=0.45\linewidth]{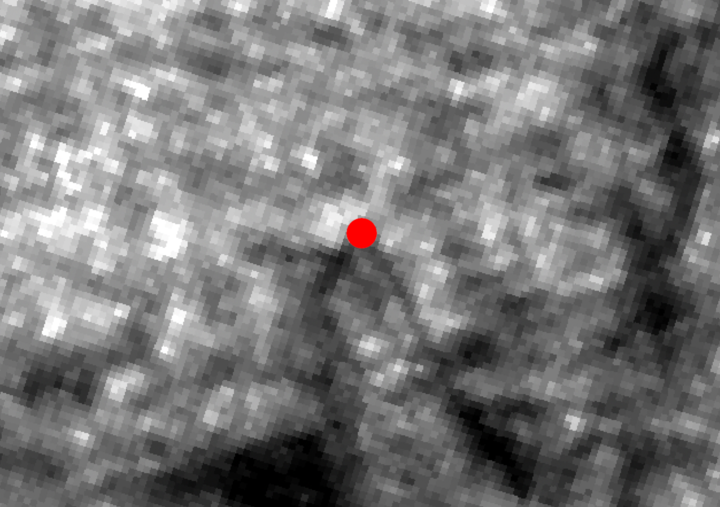} \\

        \includegraphics[width=0.45\linewidth]{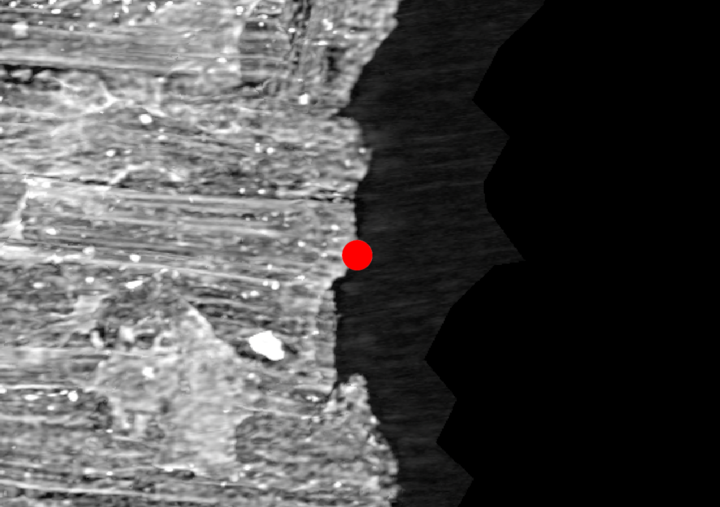} &
        \includegraphics[width=0.45\linewidth]{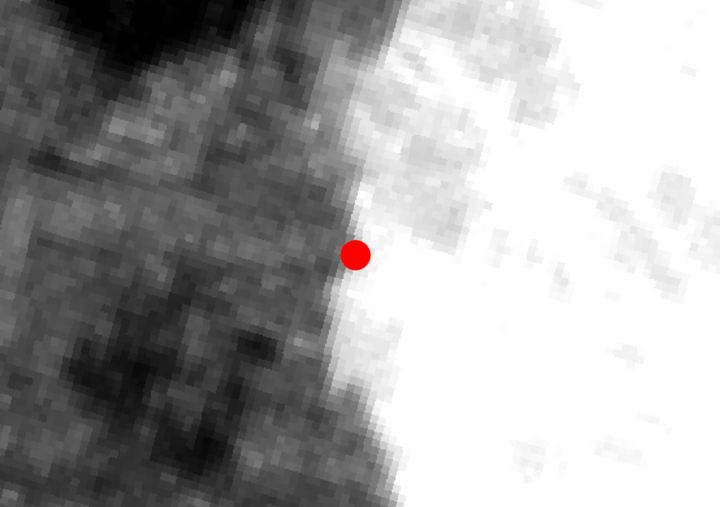} \\

        \includegraphics[width=0.45\linewidth]{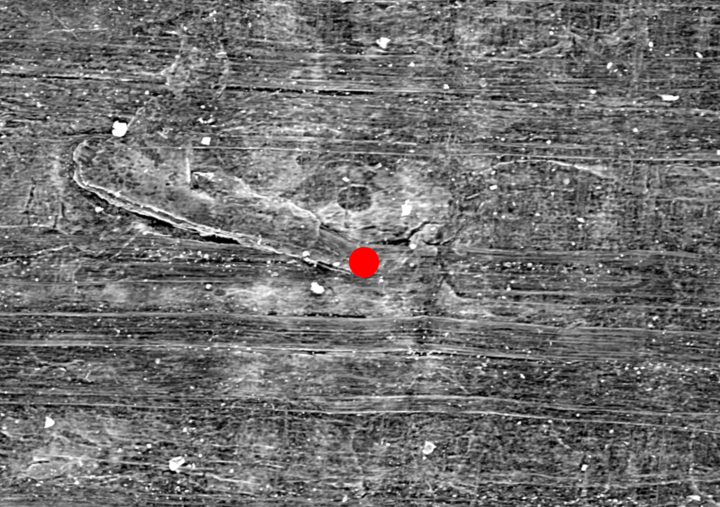} &
        \includegraphics[width=0.45\linewidth]{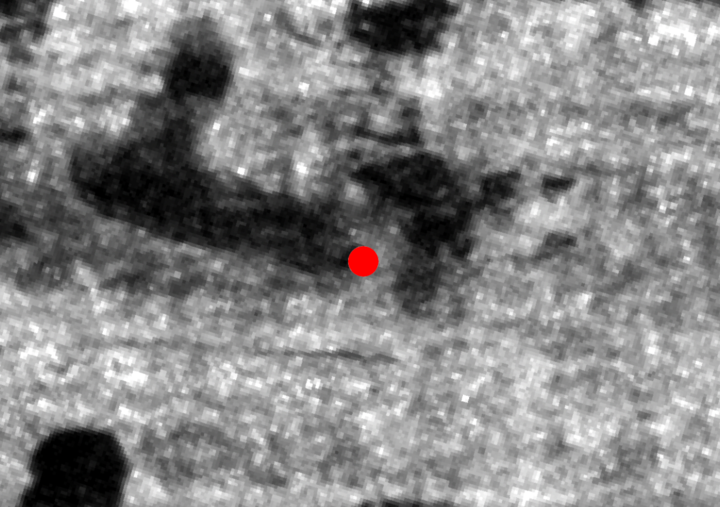} \\

        (a) Texture image. & 
        (b) Infrared. \\
    \end{tabular}
    \caption{
        Details of some example alignment points.
        Contrast stretched to enhance details.
    }
    \label{fig:alignment-detail}
\end{figure}

Instead, we align these images manually by identifying visual feature points that are common to both images.
We use Photoshop's Puppet Warp, which allows the user to specify an arbitrary number of points and their desired locations, warping the image using a deformable mesh in order to align all reference points.
Figure \ref{fig:alignment} shows the reference points used when aligning P.Herc.Paris.\ 2 fr.\ 47.
Figure \ref{fig:alignment-detail} shows details of some of these points, illustrating the sorts of features that are commonly used.

Manual feature point matching is feasible for a dataset of this scale, but is time consuming and introduces human error.
An automated method would be preferable.
Though the difference in modalities is significant and an automated method has not yet been developed for this registration problem, we believe it should be possible, especially with learned methods.
Some related work has used the small crack structure on painting surfaces for registration \cite{sindel2021craquelurenet}, and a similar approach leveraging the papyrus fiber structure could be a promising direction.

It is also possible that the registration process could be improved by using a different filter when generating the texture image.
The max filter used in this work is adequate, but other filters may generate texture images with features more resembling what is visible in infrared.
This area has not been explored.

\subsection{Label image variants}

Following alignment, the CT surface volume (Figure \ref{fig:overview-d}) and aligned infrared photograph (Figure \ref{fig:overview-f}) now comprise a dataset suitable for supervised machine learning, specifically paired image-to-image translation.
For binary ink classification, an additional step is included, where a binary segmentation mask is created by tracing the ink regions.
This is performed manually using the Quick Selection tool in Photoshop, under the supervision of a papyrologist to help disambiguate difficult spots based on papyrological context.
The resulting binary image is shown in Figure \ref{fig:overview-g}.

It should be noted that the creation of a binary mask introduces some labeling error, as it reduces the complex distribution of ink on the surface to a binary signal.
We nonetheless use the binary classification (or semantic segmentation) setup frequently, as it is a well established task with stable conventions for loss functions and other parameters.

\begin{figure}
    \centering
    \includegraphics[width=0.8\linewidth]{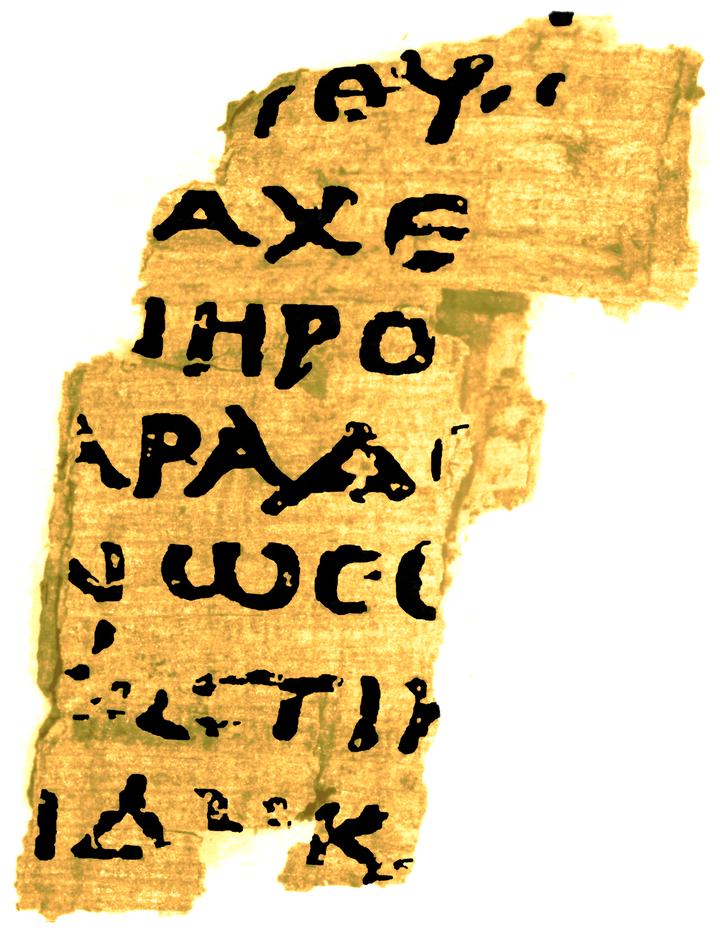}
    \caption{Stylized RGB label image generated from the infrared, illustrating possible appearance of surface before carbonization.}
    \label{fig:faux-rgb}
\end{figure}

Though not shown in Figure \ref{fig:method-overview}, we also generate an RGB image resembling what the surface may have looked like before carbonization (Figure \ref{fig:faux-rgb}).
This image is generated by color adjustments to the infrared image, using Photoshop.
Using these RGB labels, models can learn to generate images from CT inputs that resemble original, undamaged papyrus sheets.

This concludes the data preparation, and the resulting image stack is now ready for ink detection model training.

\subsection{Model training}\label{model-training}

For ink detection, the training outline and model architecture are taken from prior work \cite{parker2019inkid} with minor refinements.
The results presented here use ink-ID in binary classification mode, where the task is to view a single ``subvolume'' of input and determine whether the central voxel corresponds to ink or not.
During training and inference, points are sampled from the 2D PPM space, and are dynamically mapped to their 3D position and orientation so a subvolume can be sampled.
The subvolume is then passed as the input to a 3D convolutional neural network.

One refinement from prior work \cite{parker2019inkid} is filtering out points with ``ambiguous'' ink labels from the training set.
Since the label alignment is a manual process, some small error can be assumed.
This particularly affects the labeled pixels near the ink/no-ink boundaries, increasing the chance they are mislabeled.

\begin{figure}
    \centering
    \begin{subfigure}[b]{0.45\linewidth}
        \includegraphics[width=\linewidth]{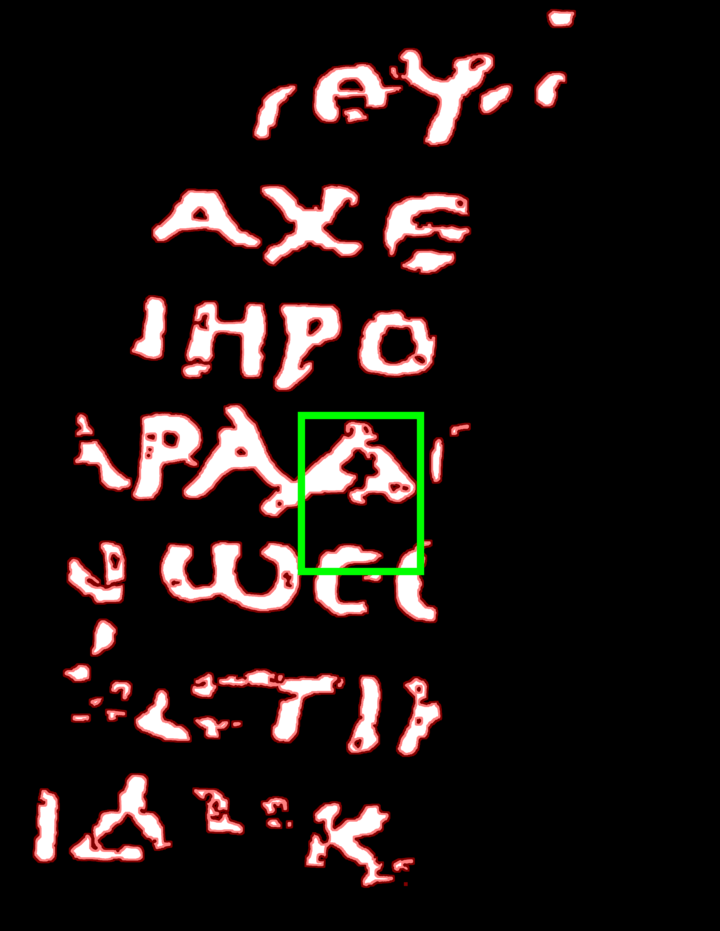}
        \caption{Entire fragment.}
    \end{subfigure}
    \begin{subfigure}[b]{0.45\linewidth}
        \includegraphics[width=\linewidth]{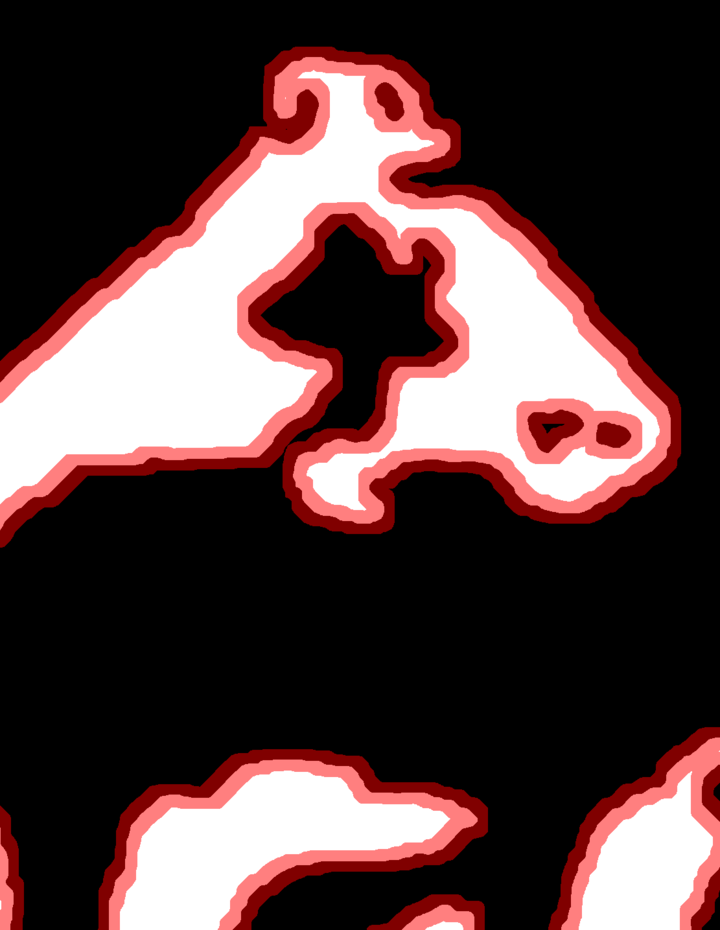}
        \caption{Detail.}
    \end{subfigure}
    \caption{``Ambiguous'' pixels (red) removed from training due to their proximity to label boundaries.}
    \label{fig:ambiguous-labels}
\end{figure}

To counter this effect, points near the ink/no-ink boundary are removed from the training set.
This exclusion region is determined by running edge detection on the ink label image, and then dilating the resulting detected edges by a configurable radius.
Figure \ref{fig:ambiguous-labels} shows the exclusion region overlaid on the ink label for P.Herc.Paris.\ 2 fr.\ 47.
This leads to improved performance in our experience, but ultimately, a more nuanced sampling or loss weighting scheme is likely preferable, such as that used in U-Net \cite{ronneberger2015u}.

The other primary refinement in the ink-ID implementation over prior work \cite{parker2019inkid} is the introduction of ``surface volumes''.
Mentioned previously, surface volumes reduce the memory footprint of the training data, allowing training across multiple fragments at once.
This increase in training dataset size and variation led to significant performance improvements.
Transitioning to surface volumes changes only the input data, and did not require changes to ink-ID.

With these refinements in mind, the fragment surface results in Figure \ref{fig:fragments-ours-surface} were created using ink-ID at commit \texttt{58b1519} with the following command:

\lstdefinestyle{mystyle}{
    basicstyle=\ttfamily\footnotesize,
    breakatwhitespace=false,
    breaklines=true,
    captionpos=b,
    keepspaces=true,
    showspaces=false,
    showstringspaces=false,
    showtabs=false,
    tabsize=2
}
\lstset{style=mystyle}
\begin{lstlisting}
inkid-train-and-predict \
    --training-set fragments_all.txt \
    --model InkClassifier3DCNN \
    --subvolume-shape-voxels 24 80 80 \
    --filters 64 64 128 64 \
    --final-prediction-on-all \
    --dataloaders-num-workers 0 \
    --checkpoint-every-n-batches 20000 \
    --prediction-grid-spacing 16 \
    --ambiguous-ink-labels-filter-radius 16 \
    --wandb \
    --output baseline_with_54keV_data \
    --cross-validate-on 0
\end{lstlisting}

Each of the four fragments was divided into top and bottom halves, so for cross validation there were eight total models trained by repeating the above command with values [0-7] for \texttt{--cross-validate-on}.
These jobs were performed in parallel on GPU nodes using the SLURM scheduling system \cite{yoo2003slurm}.
After all training jobs completed, result images were compiled using \texttt{create\_summary\_images.py} in ink-ID.
These binary prediction images were then composited using subtraction over the Volume Cartographer texture images of the respective fragments to create the images in Figure \ref{fig:fragments-ours-surface}.

\subsection{Hidden layer segmentation and inference}

Segmentation for the hidden, subsurface layers of the fragments was performed using Quick Segment (TODO citation forthcoming), a modified version of the Canny edge segmentation discussed above.
Quick Segment is a hybrid approach, using an initial manual segmentation to trace the air gap between papyrus layers, followed by a programmatic refinement step that fits the papyrus surface.

Inference was performed on the hidden layers by training a single model on all available training data, and then generating prediction images for the hidden layer surfaces.
All hyperparameters were identical to those in \ref{model-training}.
Since cross validation is not necessary, \texttt{--cross-validate-on} was omitted, and \texttt{--prediction-set} was added to specify the hidden layers.

\end{document}